\documentclass[letterpaper]{article} % DO NOT CHANGE THIS
\usepackage{aaai25}  % DO NOT CHANGE THIS
\usepackage{times}  % DO NOT CHANGE THIS
\usepackage{helvet}  % DO NOT CHANGE THIS
\usepackage{courier}  % DO NOT CHANGE THIS
\usepackage[hyphens]{url}  % DO NOT CHANGE THIS
\usepackage{graphicx} % DO NOT CHANGE THIS
\usepackage{natbib}  % DO NOT CHANGE THIS AND DO NOT ADD ANY OPTIONS TO IT
\usepackage{caption} % DO NOT CHANGE THIS AND DO NOT ADD ANY OPTIONS TO IT
\usepackage{algorithm}
\usepackage{algorithmic}
\usepackage{pifont}% http://ctan.org/pkg/pifont
\newcommand{\xmark}{\ding{55}}%
\usepackage{newfloat}
\usepackage{listings}
\DeclareCaptionStyle{ruled}{labelfont=normalfont,labelsep=colon,strut=off} % DO NOT CHANGE THIS
\floatstyle{ruled}
\newfloat{listing}{tb}{lst}{}
\floatname{listing}{Listing}
\usepackage{xr}
\usepackage{xcolor}
\usepackage{cuted}
\usepackage{subcaption}
\usepackage{multirow}
\usepackage{nccmath}
\usepackage{amsmath}

\title{EMPLACE: Self-Supervised Urban Scene Change Detection}
\author{
    Tim Alpherts,
    Sennay Ghebreab,
    Nanne van Noord
    }
\affiliations{
    University of Amsterdam\\
    t.o.l.alpherts@uva.nl, s.ghebreab@uva.nl, n.j.e.vannoord@uva.nl
}

\begin{document}

\maketitle

\begin{abstract}
Urban change is a constant process that influences the perception of neighbourhoods
and the lives of the people within them. The field of Urban Scene Change Detection (USCD) aims to capture changes in street scenes using computer vision and can help raise awareness of changes that make it possible to better understand the city and its residents. Traditionally, the field of USCD has used supervised methods with small scale datasets. This constrains methods when applied to new cities, as it requires labour-intensive labeling processes and forces a priori definitions of
relevant change. In this paper we introduce AC-1M the largest USCD
dataset by far of over 1.1M images, together with EMPLACE, a self-supervising method to train a
Vision Transformer using our adaptive triplet loss. We show EMPLACE outperforms SOTA methods both as a pre-training method for linear fine-tuning as well as a zero-shot setting. Lastly, in a case study of Amsterdam, we show that we are able to detect both small and large changes throughout the city and that changes uncovered by EMPLACE, depending on size, correlate with housing prices - which in turn is indicative of inequity. 
\end{abstract}

%     \link{Code}{github.com/Timalph/EMPLACE}

\section{Introduction}
Visual Urban Analytics (VUA) approaches have over the last decade shown potential to identify socio-economic inequity by combining computer vision techniques with street view imagery \cite{Suel2019, Naik2017}. It has been shown that visual appeal affects citizen well-being through aspects such as greenery \cite{Li2015}, perceived safety \cite{Naik2014, Ordonez2014}, or liveliness \cite{Dubey2016}, as well as more direct definitions of liveability \cite{Joglekar2020, Muller2022, Batty2019}. Proposed approaches predict either objective socio-economic metrics \cite{Suel2019,Suel2021,Law2019} or subjective perceived attributes \cite{Naik2014, Dubey2016} and show potential to support municipalities to keep a grasp on the state of their neighbourhoods \cite{Alpherts_panoparadigm_2023}. However all these approaches are static as they use a single temporal datapoint, i.e., a single image per location. As a consequence, a concept that is underexplored in VUA is the notion of urban change. Cities are constantly changing; expanding by adding new neighbourhoods and buildings, whilst existing environments decay and get renovated. Many of these changes require permits or are initiated by the municipality, which leads to awareness of these changes. However, there are also many changes to the urban environment that the municipality is not aware off - changes which can be predictive of the condition of the urban environment.

\begin{figure}[t]
\centering
\includegraphics[width=.35\textwidth]{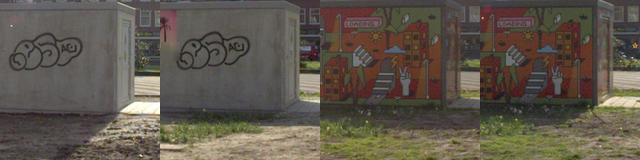}\\
\includegraphics[width=.35\textwidth]{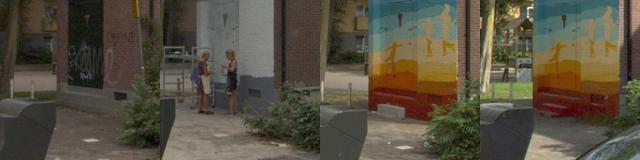}\\
\includegraphics[width=.35\textwidth]{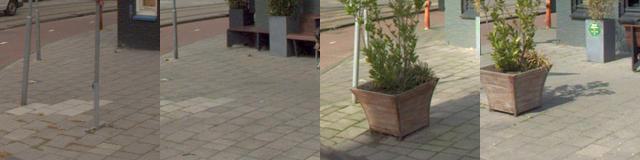}\\
\includegraphics[width=.35\textwidth]{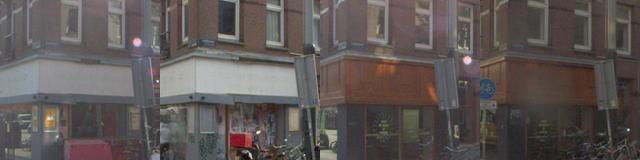}\\
\caption{Examples of various urban changes in Amsterdam uncovered through EMPLACE.}
\end{figure}

In fact, recent works within VUA have shown the relationship between urban change and socio-demographic data, by investigating how socio-economic indicators predict neighbourhood improvement \cite{Naik2017} or proposing methods to use computer vision to detect building construction \cite{huang2024citypulse}. The latter of these two approaches falls within the field of Urban Scene Change Detection (USCD), a subfield of computer vision that revolves around finding changes in sets of street view images. To further investigate the relationship between urban change and socio-demographic data we aim to extend VUA using USCD methods to explore visual changes in neighbourhoods. Unfortunately, current USCD methods are not well-equipped for this. USCD has mostly been used for specific domains such as detecting tsunami damage \cite{Sakurada2015} or autonomous driving \cite{Alcantarilla2016}, limiting their broader use within cities. Moreover, the existing datasets for USCD are relatively small, and use a variety of labelling techniques such as pixel-wise annotations \cite{Sakurada2015, Alcantarilla2016}, further limiting direct application or adaptation of existing methods. 

To study urban change from a VUA perspective it is necessary to detect change at scale throughout the entire city, whilst taking into account that the urban landscape differs tremendously between cities. This domain shift between cities is a challenge for existing USCD methods as they generally focus on \textit{discrete change}, such as a building being constructed \cite{huang2024citypulse}, which requires extensive manual labeling and an a priori definition of what is considered change. To overcome these limitations we propose a self-supervised learning approach that learns to detect urban change from unlabeled panoramic images.

To enable self-supervised learning at city-scale we built the \mbox{AC-1M} (Amsterdam Change) dataset of over 1.1M images, the largest USCD dataset by a significant margin. Furthermore, we propose EMPLACE, a self-supervised approach using an adaptive triplet loss that  learns local change features without the need for labelling procedures while being robust to fleeting changes such as cars,  people, and lighting. We evaluate its efficacy as a pre-training method for linear fine-tuning and zero-shot, and find we outperform SOTA models in both settings. By using EMPLACE we overcome the need for costly labelling processes thereby allowing our model to capture the full extent of change across the urban landscape. Our contributions are as follows:

\begin{enumerate}
    \item We build the AC-1M, the largest Urban Scene Change Detection dataset to date containing over 1.1M panoramas curated to be within a single meter of each other. %Furthermore, we also contribute two smaller datasets to evaluate discrete change: AMS-Buildings (2.3K), and AMS-Trees(1.3k).
    
    \item We introduce tEMPoraL urbAn Change lEarning (EMPLACE), a self-supervised method for learning change detection features robust to noise such as weather, people, and cars. We demonstrate EMPLACE's potential in a zero-shot setting and as a pre-training method. %Furthermore we show that EMPLACE overcomes domain shift by outperforming SOTA models on discrete change prediction as a pre-training method on both buildings and trees, as well as showing how zero-shot change prediction is a solid alternative. 

    \item In a case study of Amsterdam we show that EMPLACE can find visual elements of change without an a priori definition of what constitutes it. We uncover both small and large visual elements and show that the size of visual change correlates differently with housing prices, demonstrating that visual change is indicative of socio-economic variation across cities.
\end{enumerate}

\section{Related Work}
\subsection{Visual Urban Analytics} The field of Visual Urban Analytics has a myriad of studies that have shown that neighbourhood visuals provide insight into socio-economic indicators such as mean income \cite{Suel2019}, housing prices \cite{Law2019} or voting patterns \cite{Gebru2017}, as well as human perception such as liveliness \cite{Dubey2016}, scenicness \cite{Seresinhe2017}, uniqueness \cite{Ordonez2014}, or perceived safety \cite{Naik2014}. While the predictive capability of these methods is solid, they use black box techniques and as such cannot uncover specific visual elements that influence socio-economic inequity \cite{Alpherts_panoparadigm_2023}. More useful examples such as trash detection \cite{Sukel2020}, pothole detection \cite{Ma2022,Koch2011}, or quantifying greenery \cite{Seiferling2017} exist, but they are supervised: Using a predefined notion of what is considered an important element. By approaching VUA through USCD we could get closer to uncovering new visual elements that potentially influence the condition of a neighbourhood.

\subsection{Urban Scene Change Detection} 
The current field of USCD revolves around a supervised and \textit{bi-temporal} paradigm: datasets consist of \textit{two} images per location, taken before and after a change, which have been labelled with a segmentation map where change has taken place \cite{Varghese2018,Zhao2019, chen2021drtanet,Lei2021}. Commonly used datasets include: The PCD, consisting of 200 images of tsunami-damaged areas in japan, and the VL-CMU-CD \cite{Alcantarilla2016}, consisting of 1362 images used in autonomous vehicle navigation. Both are too narrow in scope and too small for self-supervised learning. 

Furthermore the PCD, or its successor the PSCD \cite{Sakurada2018}, are bi-temporal with aligned panoramas. In a real-world environment, over multiple years, panoramas are often misaligned due to in-the-wild differences in driving patterns of the capturing vehicle. As we are sampling our images from a real world distribution, our panorama distribution also has this misalignment requiring our method to be robust to the resulting visual noise. Finally, while panorama datasets for Amsterdam have been created before, they only consist of a single image per location and as such do not fit our task \cite{ibrahimi2021inside, yildiz2022amstertime}. 

The CityPulse dataset is the only USCD dataset with multiple images per location consisting of 4465 square viewpoint images of places with building permits. However, this dataset is still bi-temporal and supervised: consisting of pairs of images with a binary label for whether a building has been constructed or not. We refer to this notion of binary prediction as \textit{discrete} change prediction.

Attempts at steering away from supervision exist: Weakly supervised training has been employed in \cite{Sakurada2018} but this is an extrapolation of an existing labelled dataset. Self-supervised pre-training has been employed by \cite{Ramkumar2022} but revolves around training on augmented images from a supervised dataset. 
We contribute a large scale (1.1M) unsupervised dataset with multiple images per location, as well as a self-supervised method for learning change features without labels.

\section{Method}

Our goal is to learn to detect arbitrary change between two images taken at the same location in an unsupervised manner. We hypothesize that when presented with three images taken at the same location, on average, images closer in time will exhibit less change than images further away. As such we introduce EMPLACE, a self-supervised learning method for learning visual representations using an adaptive triplet loss. In the following, we describe the construction of the training dataset, the mining of triplets, and the adaptive triplet loss. Lastly we will also describe our method for evaluating our model's capability to detect change.

\subsection{Construction of the AC-1M Dataset
}

\begin{figure}
\begin{tabular}{c}
     \includegraphics[width=.45\textwidth]{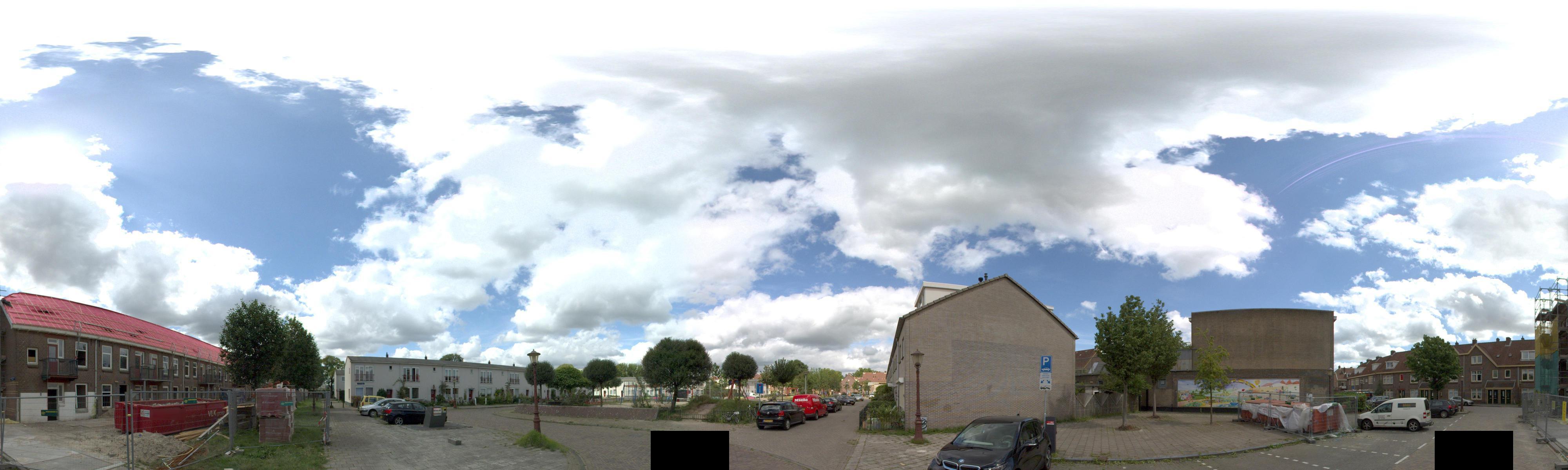} \\ 
     \includegraphics[width=.45\textwidth]{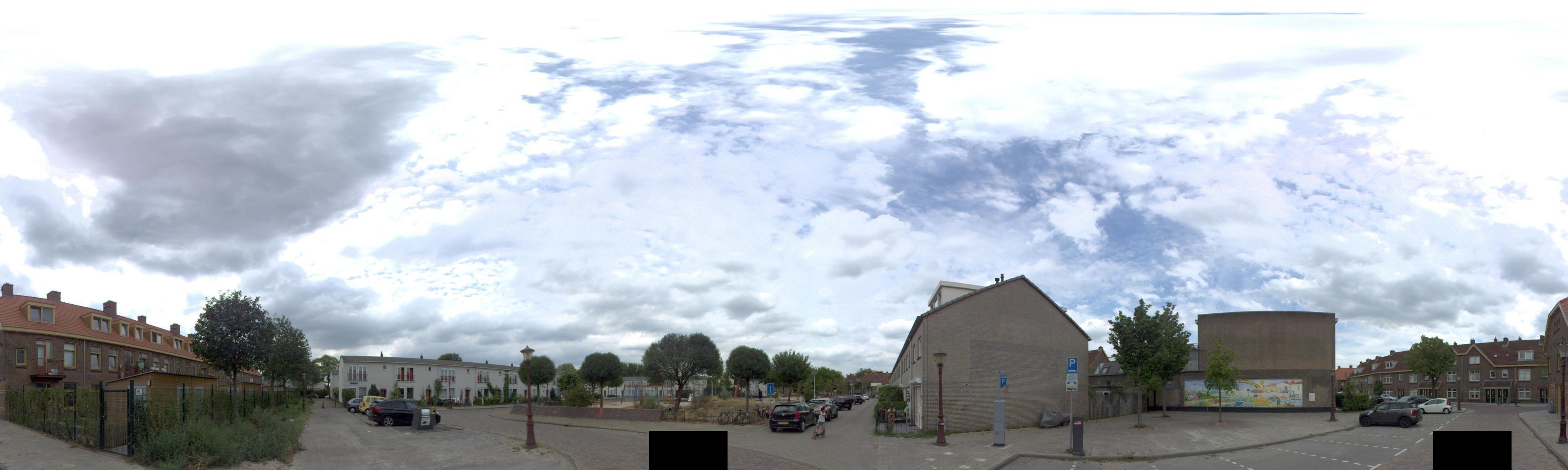}  \\ 
     \includegraphics[width=.45\textwidth]{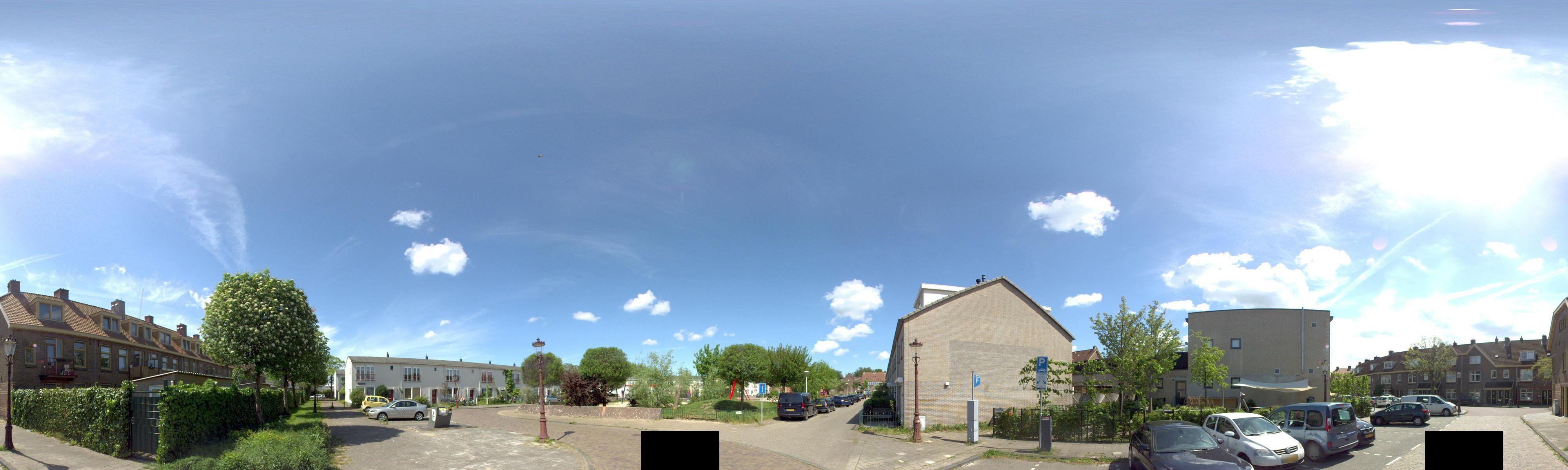} \\
\end{tabular}
 \caption{Example of a cluster from the AC-1M. From top to bottom images taken on 08-08-2016, 30-07-2018, 12-05-2022. Subtle changes happen over time such as the redoing of the roof on the left, the building of the fence on the right, and the chopping of the tree in the middle.} 
\label{fig:cluster_example}
\end{figure}

Given a time series at location $i$ of $n$ panoramic images $p^{(i)} = (p_1^{(i)}, p_2^{(i)}, ..., p_n^{(i)})$ such that $p_k^{(i)}$ corresponds to timestamp $t_k^{(i)}$, we aim to capture change in a self-supervised manner following the principle that on average $p_a^{(i)}$ and $p_b^{(i)}$ should show less change than $p_a^{(i)}$ and $p_c^{(i)}$ for $t_b^{(i)} - t_a^{(i)} < t_c^{(i)} - t_a^{(i)}$. As such we build a dataset of clusters of three or more panoramas taken at the same location. Currently, no large-scale tri-temporal change detection dataset exists for this purpose, thus we created a new dataset called the AC-1M: A dataset of panorama clusters taken within a metre of each other, we show an example in Figure~\ref{fig:cluster_example}.

The AC-1M is constructed through a clustering and selection procedure applied to the Amsterdam Panorama Database: a collection of 6M panoramas taken in Amsterdam from 2016 to 2022. The steps taken to extract clusters are as follows:
\begin{enumerate}
    \item The city is divided into polygons of its 386 neighbourhoods to reduce the computational complexity of clustering. All polygons are dilated by five metres to ensure clusters on the polygon edge are included.
    \item For each polygon, the GPS coordinates of all panoramas within it are retrieved and candidate clusters are calculated using the DBSCAN \cite{dbscan} algorithm with a radius of 1 metre and the minimum amount of samples for a cluster to be considered set at $n=3$.
    \item For each candidate cluster, the cluster centre is used to retrieve all panoramas within a radius of 1 metre. This to ensure no images appear in multiple clusters. Clusters on water are excluded as well as clusters with multiple height values (due to tunnels).
    \item Post-processing is applied by discarding the bottom 800 pixels to remove the capture vehicle. This results in panoramic images of 4000 x 1200. The heading parameter is then used to rotate all panoramas in the same direction after which a black rectangle is placed to block out the car antenna at the front and back of the vehicle, as can be seen in Figure~\ref{fig:cluster_example}. As the capturing vehicle was replaced over the years the antenna may otherwise induce a spurious correlation when predicting change. 
\end{enumerate}

\noindent The initial collection of 6M panoramas is reduced to \textbf{1.1M} panoramic images of 4000 x 1200 pixels assigned to \textbf{254k} clusters forming the \textbf{AC-1M}. This is an order of magnitude larger than existing datasets for USCD ($\leq$5000 images). A visualisation of the clusters plotted on a map of Amsterdam is shown in Figure~A.1 in the Appendix. A comparison of datasets is shown in Table~\ref{tab:datasetcomparison}. The average image per cluster is 4.29 and the median images per cluster is 4. Both the polygons and the panoramas are available from the Amsterdam Municipality API. The details on how to retrieve the AC-1M will be available at github.com/Timalph/EMPLACE.

\subsection{Triplet Mining and Loss}

To train on the AC-1M we use a triplet loss in combination with three images: an anchor, a positive, and a negative image, where the anchor and positive image are closer together in time than the anchor and negative image. Note that positive and negative describe the distance in time to the anchor, and not whether labelled change is present in the images. This allows the model to learn visual change in a self-supervised way, relying on the temporal difference as the training signal. We mine triplets from our set of clusters where a triplet consists of three images where: 
\begin{table}
    \centering
    \begin{tabular}{c|cc}
    \hline
          Dataset & $\#$ Images & $\#$ Locations\\% & Radius& Image Type\\
          \hline
         TSUNAMI &200 & 100 \\%& $<$1m&Panorama\\ 
         PSCD & 1540 & 770 \\%& $<$1m&Panorama\\
         CityPulse & 4465 & 371 \\%&$<$10m&Viewpoint\\
         AC-1M (Ours) & \textbf{1087047} & \textbf{254911} \\%&$<$1m&Panorama\\
         AMS-Buildings (Ours) & \textbf{2354} & \textbf{404} \\%&$<$1m&Viewpoint\\
         AMS-Trees (Ours) & \textbf{1374} & \textbf{273} \\%&$<$1m&Viewpoint\\
         \hline
    \end{tabular}
    \caption{Comparison of USCD datasets.}
    \label{tab:datasetcomparison}
%\end{subtable}
\end{table}

\begin{align}
  \Delta_{AP} &< \Delta_{AN} \\
  \text{where}~ \Delta_{AP} &= t_{pos}^{(i)} - t_{anc}^{(i)} \\
  \Delta_{AN} &= t_{neg}^{(i)} - t_{anc}^{(i)}\\
  \Delta_{PN} &= t_{neg}^{(i)} - t_{pos}^{(i)}
\end{align}
which results in the amount of triplets that can be mined from a cluster of size n being $\binom{n}{3}$. 
Note that for all triplets:
$$t_{anc}^{(i)}<t_{pos}^{(i)}<t_{neg}^{(i)}$$
The full set of possible mined triplets is 2 million. For our purposes these triplets are filtered along certain constraints to steer them towards capturing meaningful changes. For example, by restricting the anchor and positive image to be within thirty days of each other, and the anchor and negative image to be more than three years apart. Note that this method allows the model to naturally become robust to changes in weather and lighting conditions.

To train using these triplets we use an adaptive triplet loss as shown in Eq.~\ref{eq:vanilla_triplet_loss} where the margin $\alpha$ is defined in Eq.~\ref{eq:temporal_triplet_loss}. 
We use a scaling margin as the observed changes are not linear with respect to time, as sudden changes can occur in between images. For images taken close together we often find that there is no change, whereas if there are multiple years in between images there may be many large changes. Yet, this process of change is typically not gradual but rather happened as a sudden jump in between two images.
 
\begin{align}
    \mathcal{L}(p^{(i)}_{anc},p^{(i)}_{pos},p^{(i)}_{neg}) = max(  ||f(p^{(i)}_{pos}) - f(p^{(i)}_{anc})||_2 \notag\\ -  ||f(p^{(i)}_{neg}) - f(p^{(i)}_{anc})||_2 + \alpha, 0) \label{eq:vanilla_triplet_loss}\\ 
    \alpha = 
    \begin{cases}
      0.5 * \left(\frac{\Delta_{PN}}{365}\right) ^2 & \text{if } \Delta_{PN} < 365 \\\\
      \frac {\Delta_{PN}}{365} - 0.5 & \text{else} 
    \end{cases}
    \label{eq:temporal_triplet_loss}
\end{align}

\subsection{Model}
\begin{figure}
    \centering
    \includegraphics[width=0.48\textwidth]{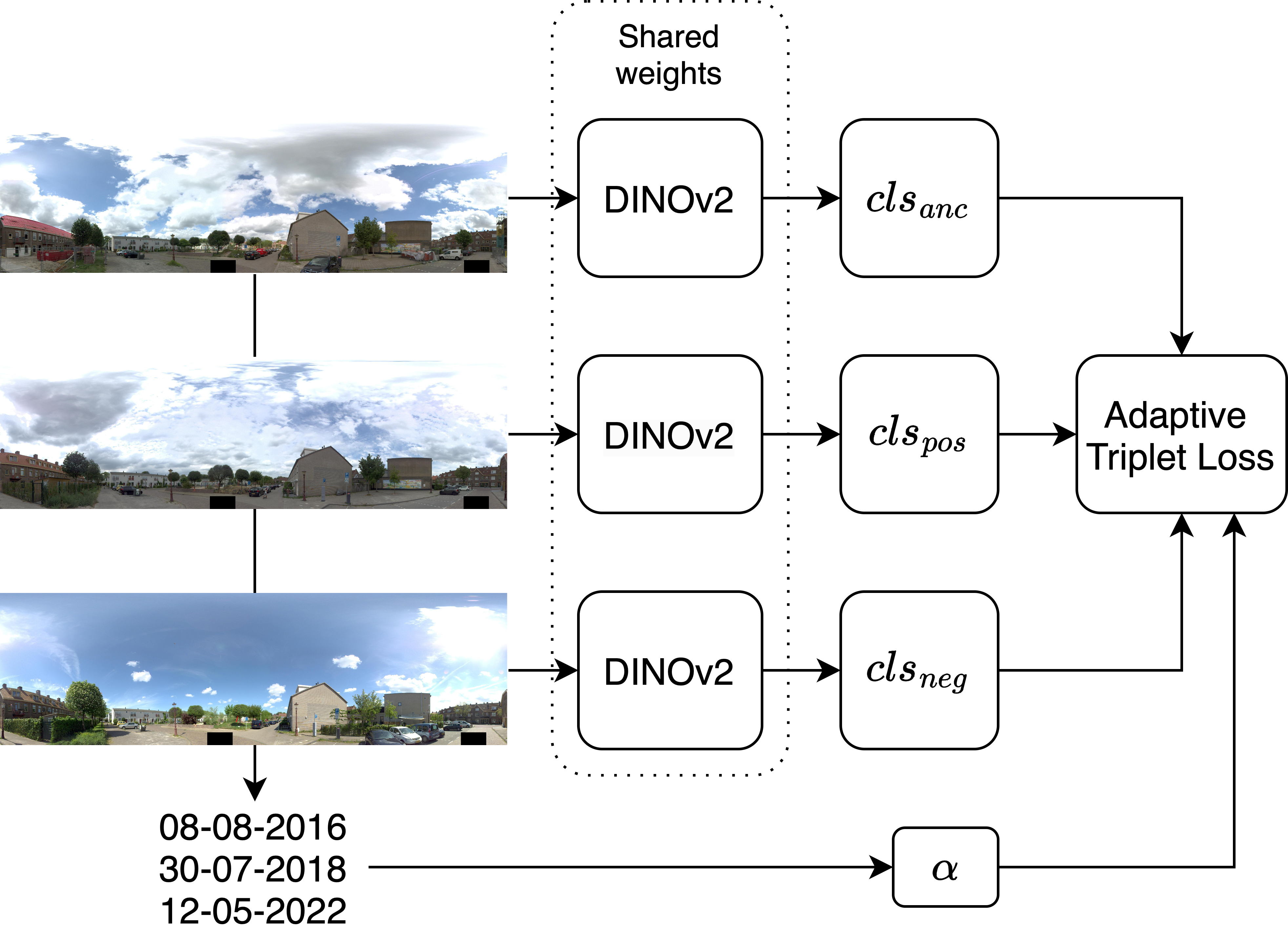}
    \caption{Overview of the model architecture. Image triplets perform a forward pass through a Siamese backbone with DINOv2 weights to calculate the $cls$ tokens. The image dates are used to calculate the margin $\alpha$. The $cls$ tokens and margin $\alpha$ are used to calculate the adaptive triplet loss.}
    \label{fig:modelflowarchitecture}
\end{figure}

For our transformer model we use ViT-B/14 initialised with DINOv2 weights \cite{oquab2024dinov2} as it has been shown to outperform other methods by a comfortable margin on the task of USCD \cite{huang2024citypulse}. Our images are downsized to 700x210 for analysis, which results in the use of 14x14 patches that divide the image into a sequence length of 751, including the $cls$ token. An overview of the model architecture is shown in Fig~\ref{fig:modelflowarchitecture}.

\begin{figure}
\centering
\includegraphics[width=.46\textwidth]{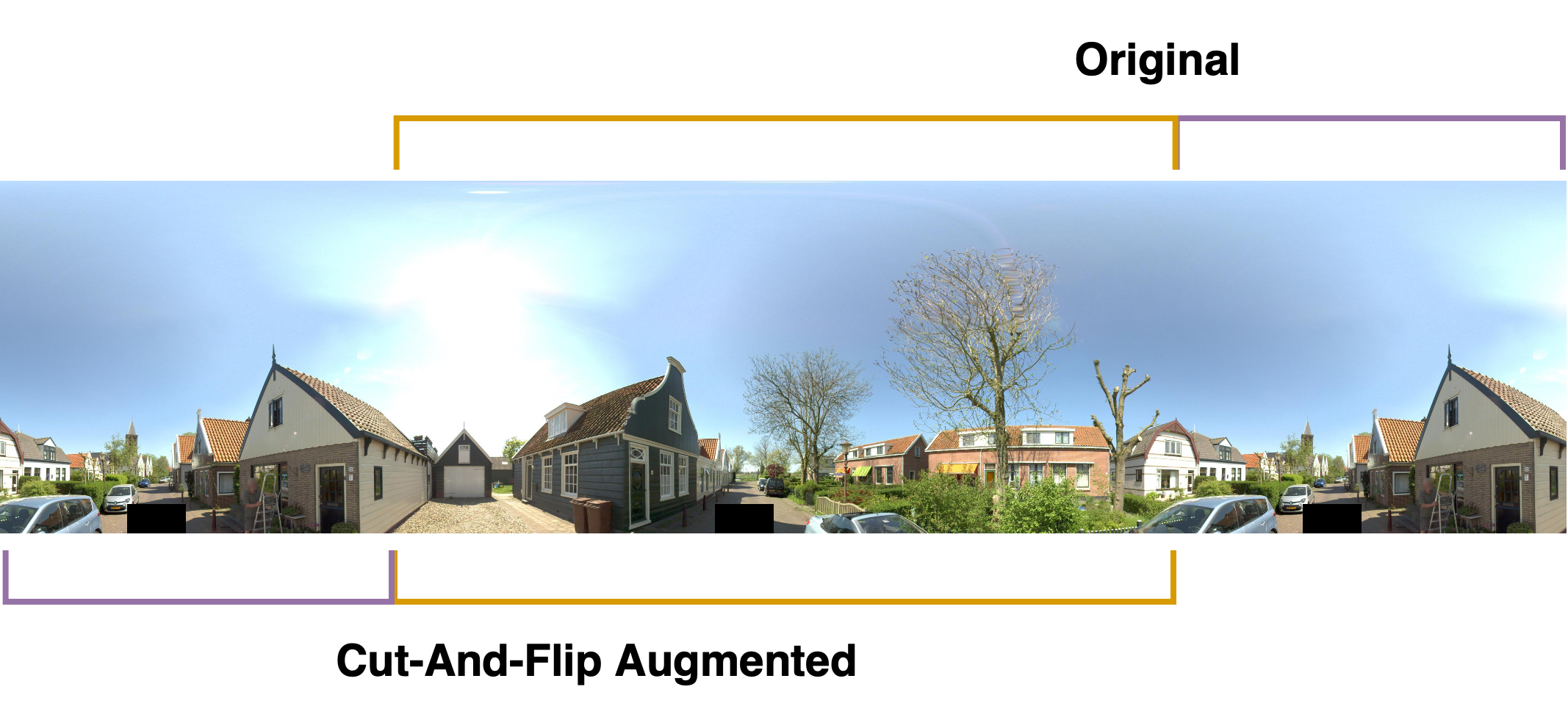} 
\caption{An example of cut-and-flip data augmentation.} 
\label{fig:cutandflip}
\end{figure}

During training we introduce a new data augmentation step we call cut-and-flip augmentation: Panoramas differ from regular images in that they are circular. They consist of images taken in four directions of a capturing vehicle that are projected to a visual space where the horizontal axis of the image is circular and periodic.
We enforce this visual nature through cut-and-flip augmentation where given a triplet, a random vertical cut is made and the cut images are swapped around. An example of cut-and-flip is shown in Figure~\ref{fig:cutandflip}.

To conclude, each training iteration consists of: 1) a forward pass of a triplet where the triplet loss is calculated applied to the euclidean distances of the resulting $cls$ tokens, 2) a forward pass with the cut-and-flip augmented triplet, and 3) a backward pass performed using the cumulative loss. 

\section{Experiments and Evaluation}

In this section we will describe our procedure for training on the AC-1M, the parameters, and our evaluation procedure through order prediction. We will then describe our method for discrete change prediction, the models we evaluate against, and the creation of two \textit{new} discrete change detection datasets: AMS-Buildings and AMS-Trees.
\begin{table}
    \centering
    \begin{tabular}{c|c|c||c}
              &  $\Delta_{AP}$&$\Delta_{AN}$&$\#$ of triplets\\
              \hline
         SI-1 &  $1<x<31$&$375<x$&   $14361$\\
         SI-2 &  $275<x<475$&$750<x$&  $77125$\\
         SI-3 &  $275<x<475$&$1125<x$& $36384$\\
         SI-4 &  $275<x<475$&$1500<x$& $13344$\\
         \hline
    \end{tabular}
    \caption{Distances in days between the anchor and positive image, and anchor and negative image for different SI setups.}
    \label{tab:SI-table}
\end{table}

\section{Training EMPLACE}

We randomly split the AC-1M dataset into a training, validation, and test set using a 70/20/10 split. This split is performed by cluster, which means every cluster only appears in one set. We decide on triplet constraints by calculating the SI, or Sampling Interval, which is 375 days. This corresponds to the median time between images in clusters of the AC-1M. We use the SI to evaluate four training setups: SI-1, SI-2, SI-3, SI-4 where the number is used to describe $\Delta_{AN}$ in SI. For our setups $\Delta_{AP}$ is between 275 and 475 days, except for SI-1 in which it is between 1 and 31 days. The lower bound of $\Delta_{AP}$ is always more than 1 day to ensure the task does not become too easy. These training setups, alongside the number of triplets after filtering are shown in Table~\ref{tab:SI-table}. For each SI setup we train a ViT-B/14 using the adaptive triplet loss and cut-and-flip augmentation. We use the Adam optimizer with a learning rate set to $10^{-5}$, a batch size of 64, and grad norm set to $\leq0.5$. Training and evaluation was conducted on 4 NVIDIA GeForce GTX 1080 Ti GPUs. 
\subsection{Order Prediction} To evaluate model accuracy and capability to detect change on the validation and test set we introduce the task of \textit{order prediction}. Where the model is presented with a triplet and has to place the positive and negative image in the right temporal order based on euclidean distance to the anchor. 
We utilize early stopping if the accuracy has not improved for five epochs. While the validation set consists only of triplets using the same SI setup as the training set, after training we also evaluate every SI training setup on the test sets of other SI setups as well as on all test sets combined. 

\subsection{Evaluating Discrete Change} After training our EMPLACE model in a self-supervised manner, we also evaluate the performance of EMPLACE on \textit{discrete} change prediction as introduced by \cite{huang2024citypulse}. In this setting the model is presented with two viewpoint images and is tasked to predict whether change has occurred. This task is \textit{supervised} and as such we will  describe the construction of two labelled datasets for this purpose, the linear head necessary to turn the cls prediction into a discrete output, fine-tuning setup, training parameters, and backbones we test alongside EMPLACE.  

\subsection{AMS-buildings and AMS-Trees} 
\begin{figure}
    \centering
    % First subfigure
    \begin{subfigure}[b]{0.22\textwidth}
        \centering
        \includegraphics[width=\textwidth]{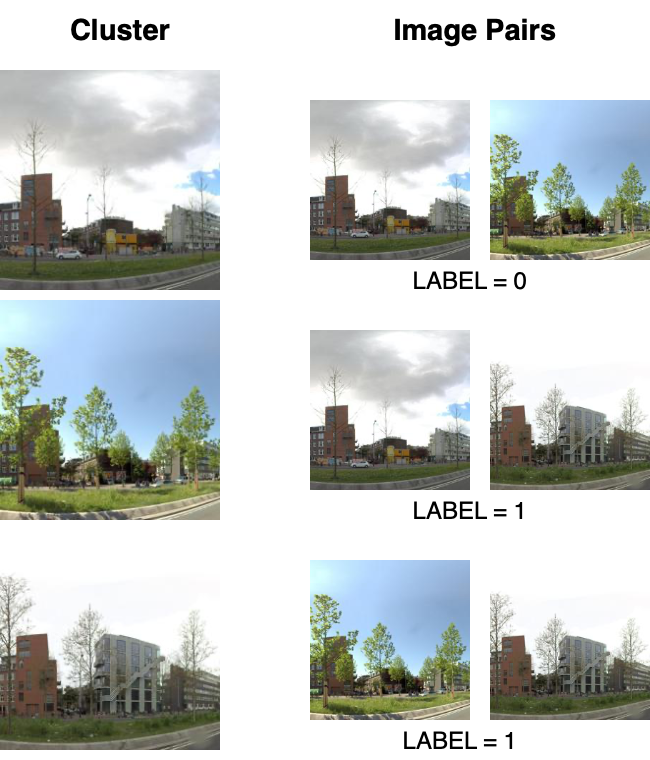}
        \caption{AMS-Buildings}
        \label{fig:subfig1}
    \end{subfigure}
    \hfill
    % Second subfigure
    \begin{subfigure}[b]{0.22\textwidth}
        \centering
        \includegraphics[width=\textwidth]{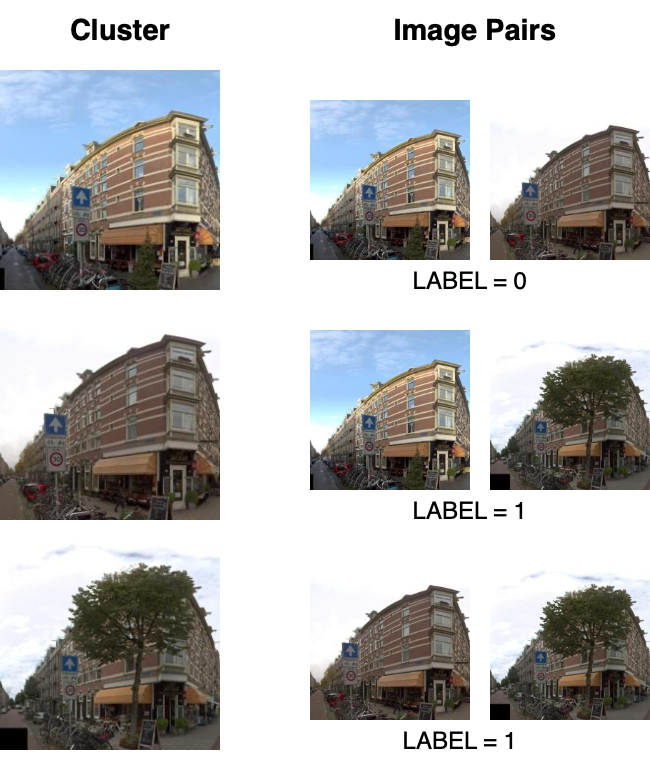}
        \caption{AMS-Trees}
        \label{fig:subfig2}
    \end{subfigure}
    \caption{Examples of cluster and image pairs from AMS-Trees and AMS-Buildings.}
    \label{fig:AMS-TREES-BUILDINGS-EXAMPLE}
\end{figure}

\begin{figure}[ht]
\begin{tabular}{cc}
     \includegraphics[width=.22\textwidth]{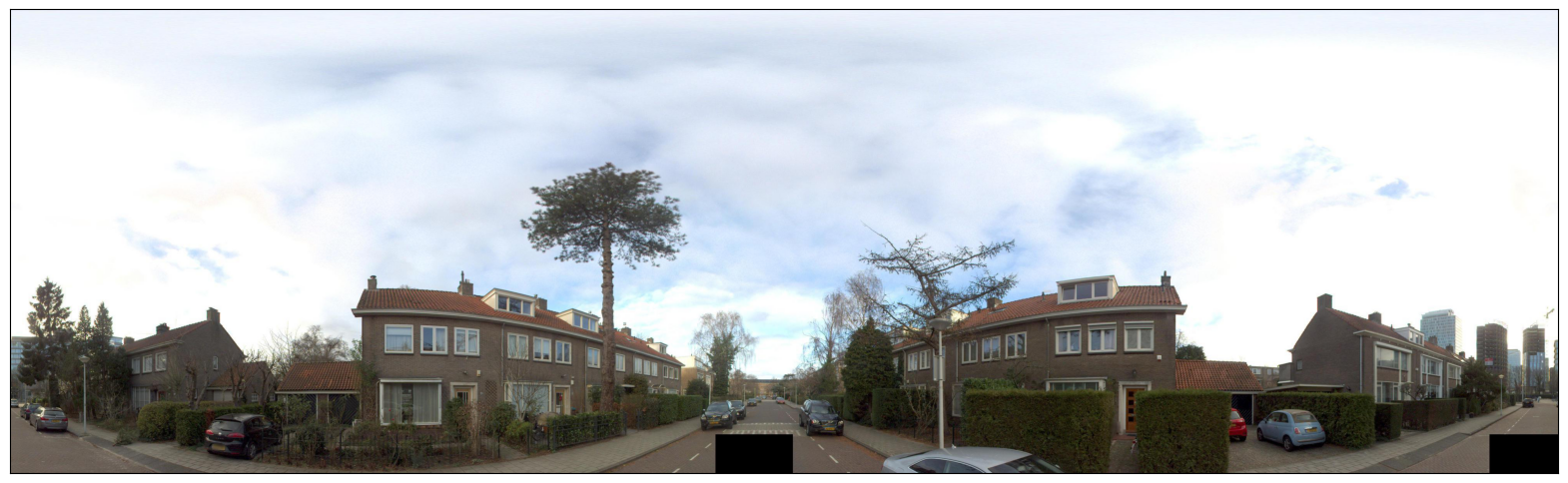}&\includegraphics[width=.22\textwidth]{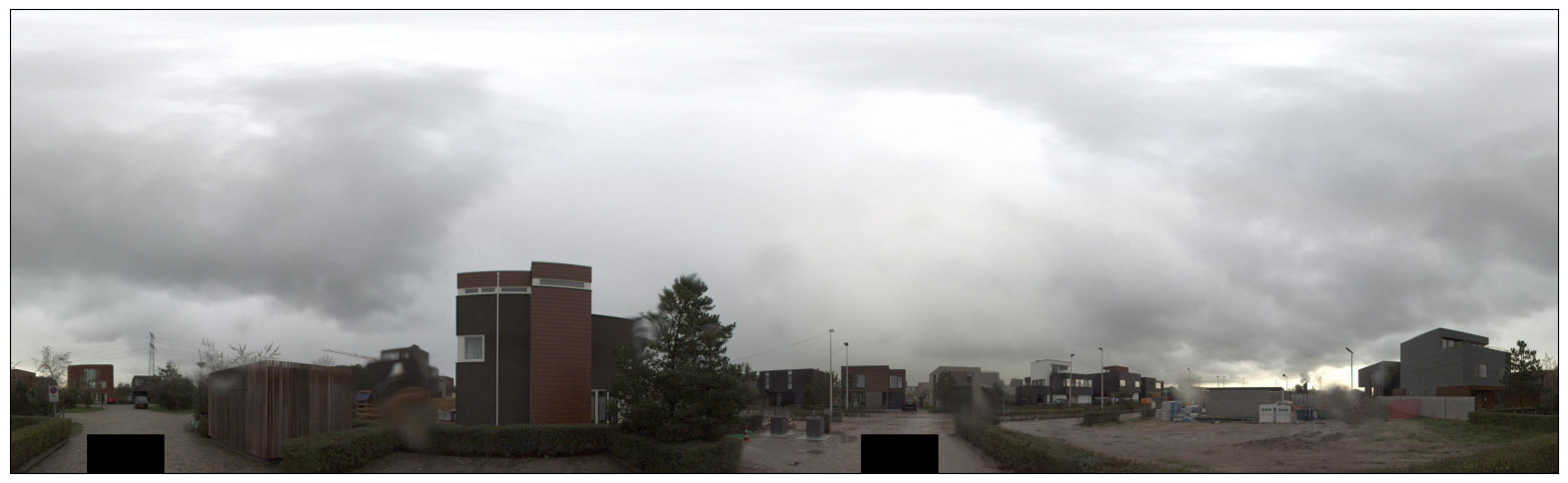} \\ 
     \includegraphics[width=.22\textwidth]{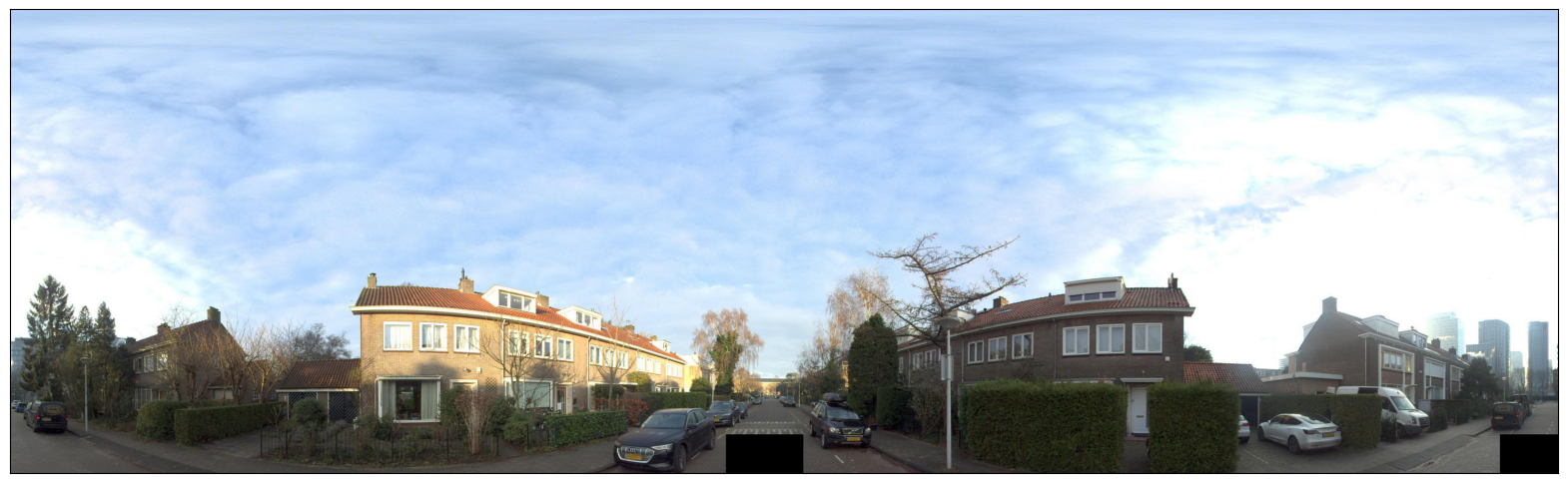}&\includegraphics[width=.22\textwidth]{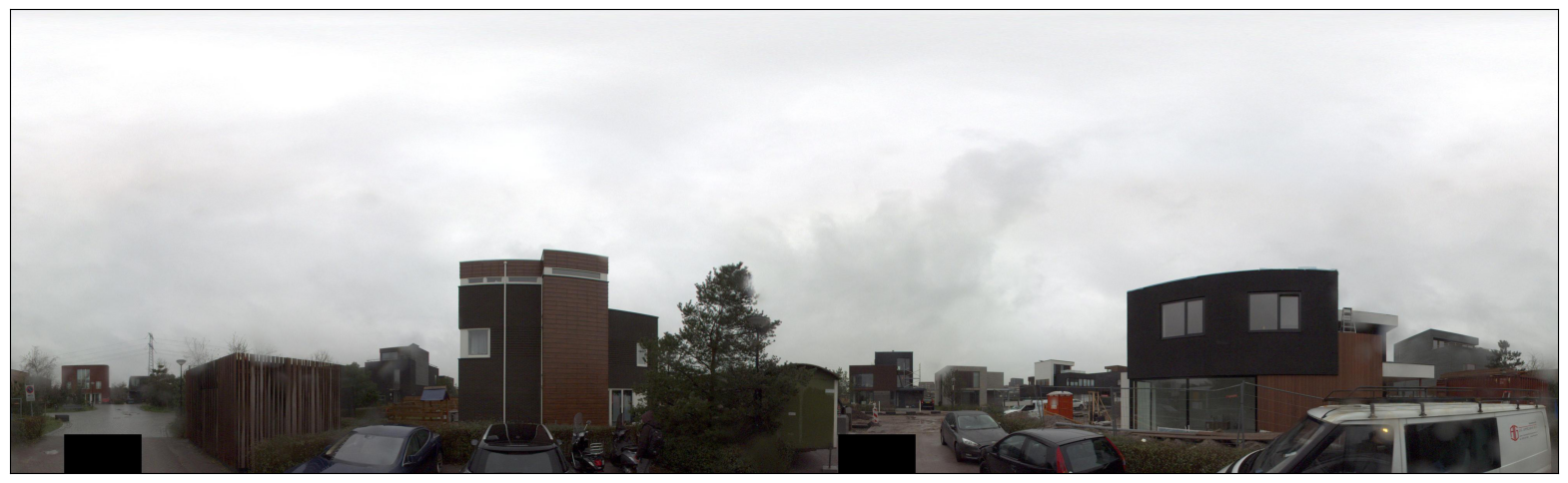}  \\ 
     \includegraphics[width=.22\textwidth]{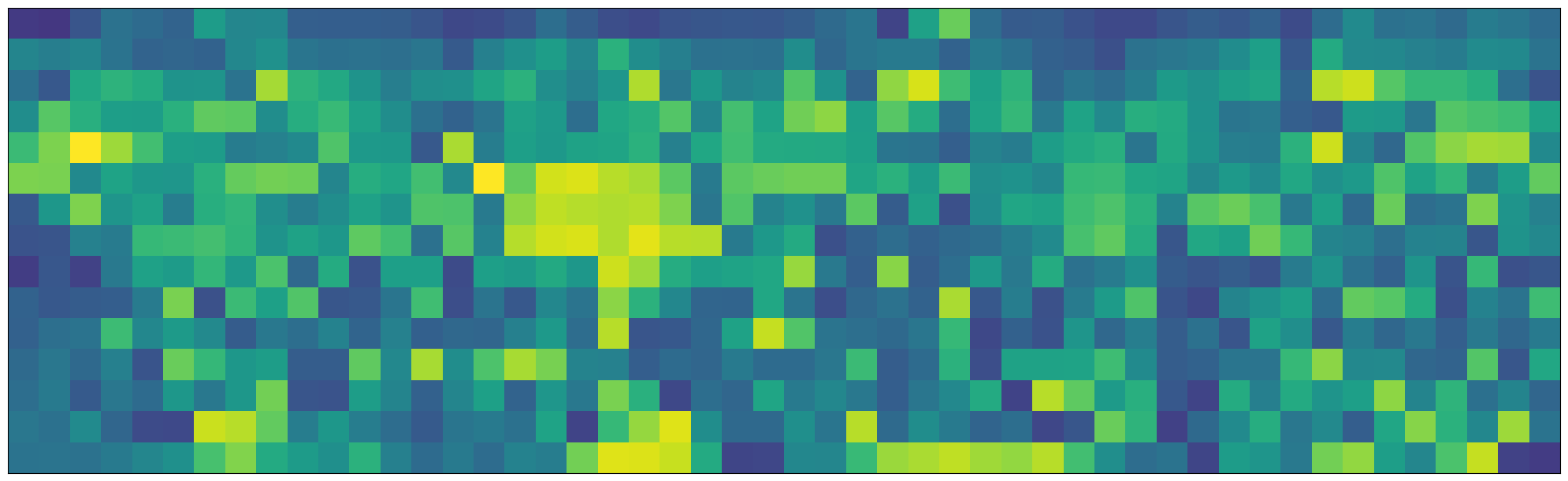}&\includegraphics[width=.22\textwidth]{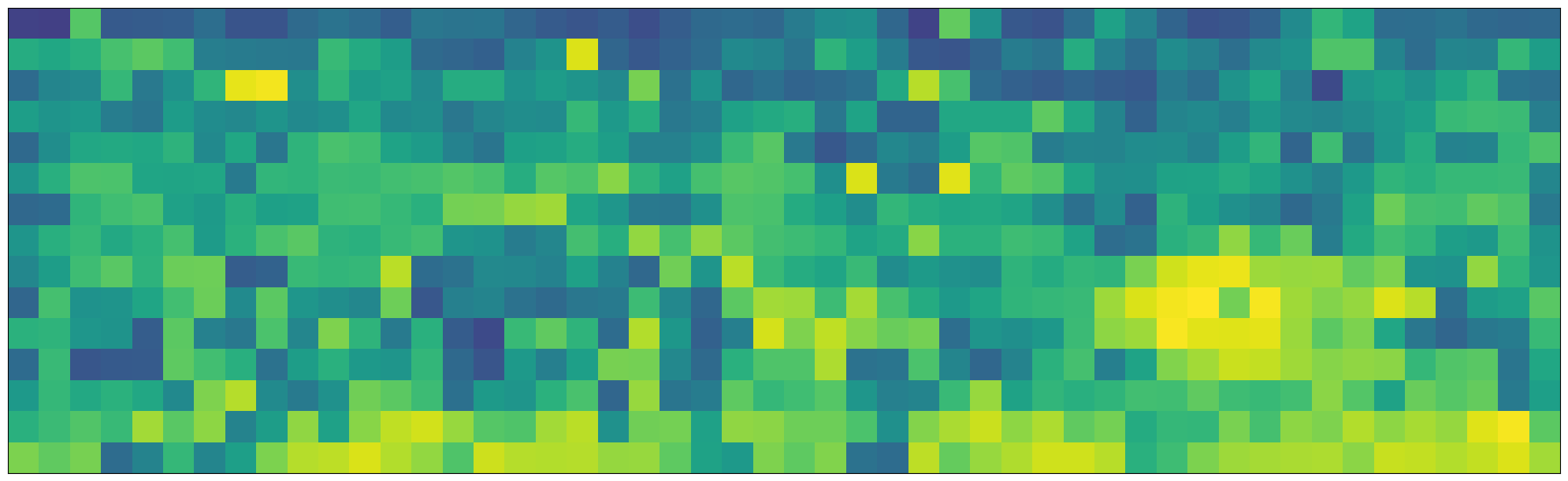} \\
     \includegraphics[width=.22\textwidth]{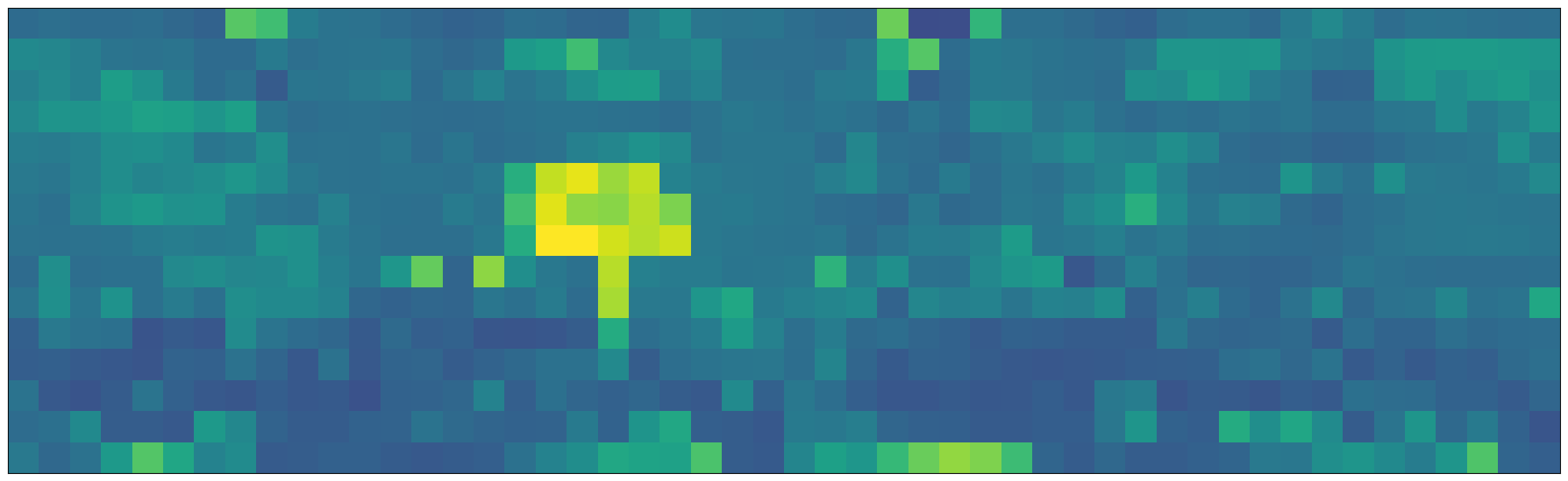}&\includegraphics[width=.22\textwidth]{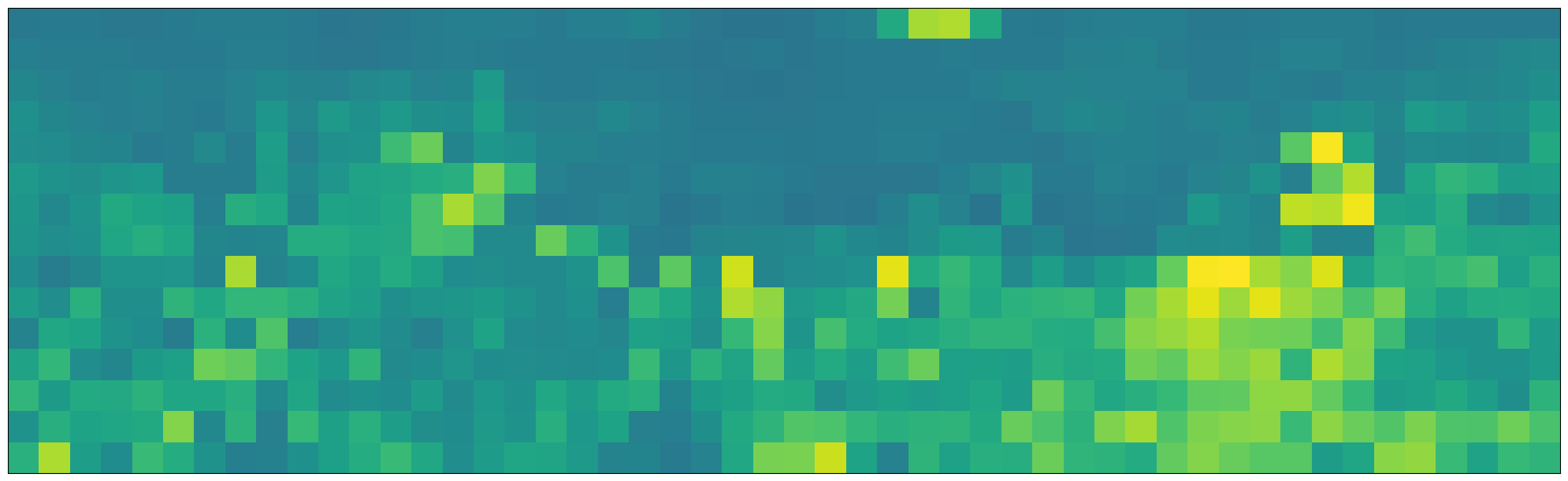} \\
     \includegraphics[width=.22\textwidth]{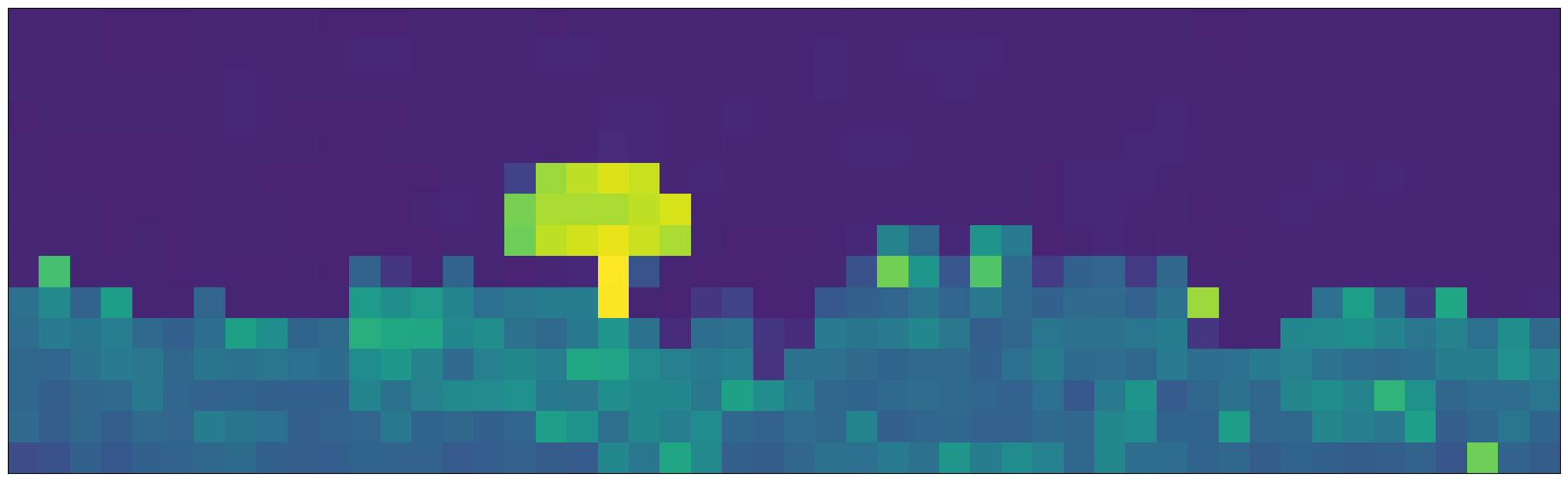}&\includegraphics[width=.22\textwidth]{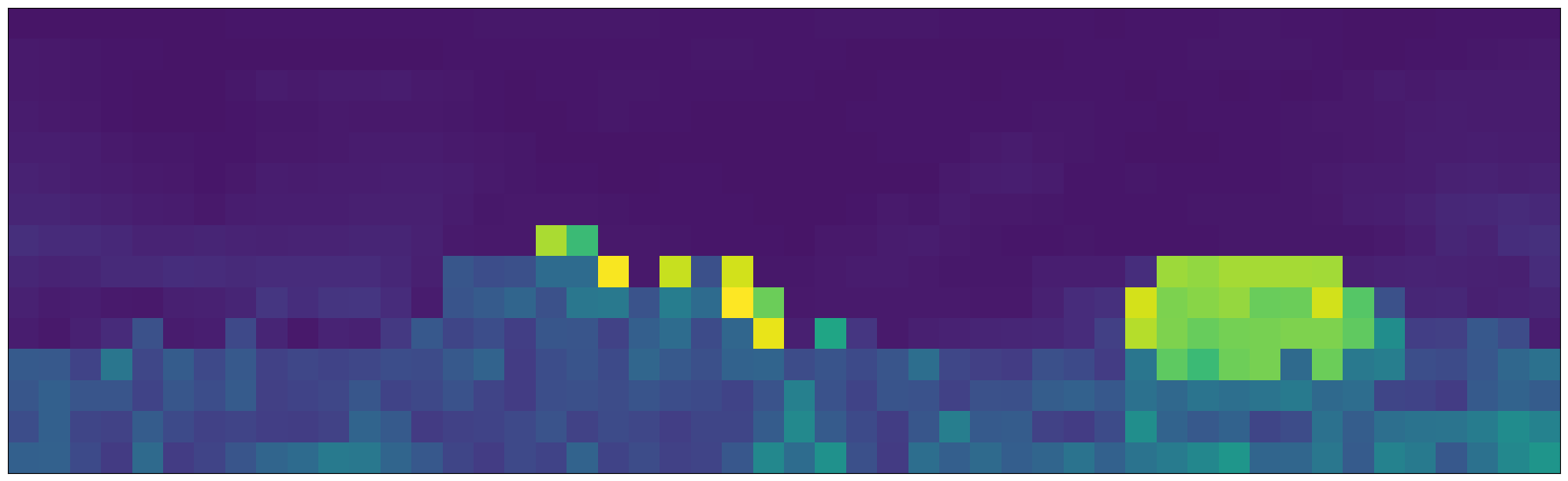} \\
\end{tabular}
\caption{Heatmap comparison. Top two rows show the images between which the heatmaps are computed. Row three is ImageNet, row four is after training for discrete change, row five is zero-shot EMPLACE.} 
\label{fig:heatmaps}
\end{figure}

\begin{table}
    \centering
    \begin{tabular}{cc|c|c|c|c|c|}
         \multicolumn{2}{c}{} & \multicolumn{5}{c}{\textbf{Test set}} \\
         \cline{3-7}
         \multirow{6}{*}{\rotatebox{90}{\textbf{Training Setup}}} & & SI-1 & SI-2 & SI-3 & SI-4 & All \\
         \cline{2-7}
         & SI-1 & \textbf{.975} & .683 & .726 & .742 & .744 \\
         & SI-2 & .959 & \textbf{.903} & \textbf{.948} & .978 & \textbf{.936} \\
         & SI-3 & .926 & .834 & .937 & \textbf{.981} & .900 \\
         & SI-4 & .895 & .771 & .846 & .952 & .840 \\
         & \xmark & .709 & .591 & .602 & .624 & .616 \\
        \cline{2-7}
    \end{tabular}
    \caption{Results for EMPLACE on the task of order prediction, with the training setup in the form of SI-X on the y-axis, and the test set on the x-axis. The last column is the cumulative score on all test sets normalized by test set size. The last row shows the performance without training, or a ViT-B/14 model with DINOv2 features.}
    \label{tab:order_prediction_table}
\end{table}
To evaluate EMPLACE on the task of discrete change prediction we construct two labelled datasets using the same method as \cite{huang2024citypulse}, the SOTA method for evaluating discrete change prediction: AMS-Buildings and AMS-Trees. Both datasets are constructed from the AC-1M test set to ensure the images are not seen by EMPLACE during training. AMS-Buildings contains $2327$ image pairs \textit{with} change and $1815$ without. AMS-Trees contains $1157$ with change, and $1084$ without. Examples of images from both datasets are visible in Figure~\ref{fig:AMS-TREES-BUILDINGS-EXAMPLE}, while comparisons to existing datasets are shown in Table~\ref{tab:datasetcomparison}. Both datasets will be available at github.com/Timalph/EMPLACE.

\begin{table*}
    \centering
    \begin{tabular}{c||c||ccccc}
          \multicolumn{5}{c}{Buildings} \\
    \hline
          Model&Pre-Training& Acc & Prec & Rec & F1\\

\hline\hline
ResNet101&\xmark& $.647\pm.013$&$.709\pm.007$&$\textbf{.851}\pm.029$&$.772\pm.011$\\
CLIP &\xmark&$.706\pm.023$&$.74\pm.027$&$.775\pm.026$&$.755\pm.019$\\
DINOv2 &\xmark&$.701\pm.023$&$.764\pm.02$&$.798\pm.027$&$.778\pm.014$\\
\textbf{EMPLACE}&SI-2&$.732\pm.031$&$.802\pm.04$&$.792\pm.031$&$\textbf{.794}\pm.016$\\
\textbf{EMPLACE zero-shot}&SI-2&$.761\pm.025$&$.792\pm.027$&$.718\pm.074$&$.75\pm.037$\\
\textbf{EMPLACE zero-shot}&SI-4&$\textbf{.781}\pm.022$&$\textbf{.840}\pm.031$&$.703\pm.058$&$.763\pm.031$\\

\hline\hline
\multicolumn{5}{c}{Trees} \\
\hline
          Model&Pre-Training& Acc & Prec & Rec & F1\\
\hline\hline
ResNet101&\xmark&$.705\pm.046$&$.600\pm.111$&$.819\pm.067$&$.676\pm.071$\\
CLIP&\xmark&$.74\pm.037$&$.686\pm.124$&$.759\pm.083$&$.7\pm.058$\\
DINOv2&\xmark&$.762\pm.028$&$.758\pm.116$&$.739\pm.088$&$.732\pm.051$\\
\textbf{EMPLACE}&SI-2&$\textbf{.855}\pm.003$&$\textbf{.853}\pm.007$&$\textbf{.924}\pm.007$&$\textbf{.885}\pm.001$\\
\textbf{EMPLACE zero-shot}&SI-2&$.765\pm.033$&$.803\pm.062$&$.752\pm.068$&$.773\pm.038$\\
\textbf{EMPLACE zero-shot}&SI-4&$.764\pm.026$&$.799\pm.029$&$.718\pm.073$&$.753\pm.036$\\

\hline\hline
    \end{tabular}
    \caption{Results on AMS-Buildings and AMS-Trees. Best scores are shown in bold.}
    \label{tab:CityPulseEval}
%\end{subtable}
\end{table*}

\noindent\textbf{Training setup} We test the performance of EMPLACE as a pre-training method on the AMS-Buildings and AMS-trees by splitting both datasets into train, validation, and test sets following a 70/20/10 split and fine-tune the EMPLACE model that scores best on order prediction to perform discrete change prediction. We use a Siamese network with EMPLACE as the twin backbone model to compute the cls tokens of an image pair and concatenate them into a final layer to transform the input into a scalar prediction as shown in Eq. 3 in \cite{huang2024citypulse}. We assess the performance of EMPLACE by comparing it to three pre-trained backbone models: ResNet101 \cite{he2015deepresiduallearningimage}, DINOv2 \cite{oquab2024dinov2}, and CLIP \cite{radford2021learningtransferablevisualmodels}. We fine-tune all models on both the AMS-Buildings and AMS-Trees. We use the Adam optimizer with a learning rate set to $10^{-5}$, a batch size of 16, grad norm set to $\leq0.5$, and early stopping after not having improved for 3 epochs. Training and evaluation was conducted on 1 NVIDIA GeForce GTX 1080 Ti GPU.  We report the means and standard deviations of 10 runs with different seeds on the test set.

\subsection{Visualising Change Detections}
As our EMPLACE model has learned features for change detection during training we can compute a heatmap by using the euclidean distance between the token output of the Vision Transformer for zero-shot change prediction. An example of these heatmaps is shown in Figure~\ref{fig:heatmaps}. Our method for zero-shot change prediction therefore consists of moving a window over the image, and calculating the mean value of the euclidean distances in the patches of the window. If anywhere on the heatmap this mean value is above a certain threshold we output a detected change, and output no detection if otherwise. The optimal threshold and window size are both learned solely on the validation set.

\section{Results}

\subsection{Order Prediction} 
Order prediction is the task of putting two images in the correct temporal order with respect to an anchor image. As such, we consider this a simple evaluation method to evaluate the change detection properties of USCD models. The results for the order prediction task are shown in Table~\ref{tab:order_prediction_table}. We see that SI-2 EMPLACE (EMPLACE trained on SI-2) scores best overall by quite a large margin. Presumably this is because SI-2 EMPLACE has the hardest setup, relatively having the closest temporal distance between positive and negative images. The performance of SI-2 EMPLACE on the SI-3 test set, as well as high scores on the SI-1 and SI-4 test sets also indicate that SI-2 EMPLACE is most robust to visual noise present in urban images, and as such has learnt visual change most effectively. We see that while SI-1 EMPLACE outperforms the other models on its own test set, this performance does not generalize to other test sets indicating that the training objective of SI-1 EMPLACE is too easy: in SI-1 EMPLACE both anchor and positive image are taken during the same season, which is not the case for the other setups. Another interesting result is that SI-4 underperforms SI-2 and SI-3 on its own test set, seemingly indicating that the training set of SI-4 suffers from being too small; The stricter our triplet constraints are, the smaller the training set will be. Lastly, we see that this task is not trivial, as without training the model scores only .616 across all test sets.

\subsection{Discrete Change Evaluation}
The results of the discrete change evaluation experiments on the AMS-Buildings and AMS-Trees are shown in Table~\ref{tab:CityPulseEval}. We see that EMPLACE pre-training outperforms vanilla backbones when fine-tuned on both AMS-Buildings and AMS-Trees. We also see that the margins are higher on the AMS-Trees dataset. We assume this is due to change regarding trees being much more present in the AC-1M, while building construction is more scarce. We also observe that EMPLACE zero-shot actually outperforms the Siamese backbone on accuracy and precision on the AMS-Buildings. Furthermore, EMPLACE SI-4 zero-shot performs better than EMPLACE SI-2, potentially indicating that a larger $\Delta_{AN}$ forces the model to rely more on the built environment. The fine-tuning results indicate that EMPLACE is able to learn domain specific features about the Amsterdam that increase its performance on change detection tasks. The zero-shot performance indicates that the change detection task is also adequately doable without supervision.
Lastly, we perform two additional studies: the first is an ablation study to evaluate the utility of the adaptive triplet loss and cut-and-flip augmentation, the results are shown in Appendix Table~A.1. The second is a study on whether our method's performance is biased towards locations that show more change, the results of this are shown in Appendix Table~A.2.

\begin{figure*}[!ht]
    \centering
    \begin{subfigure}{0.49\textwidth}
        \centering
        \includegraphics[width=\textwidth]{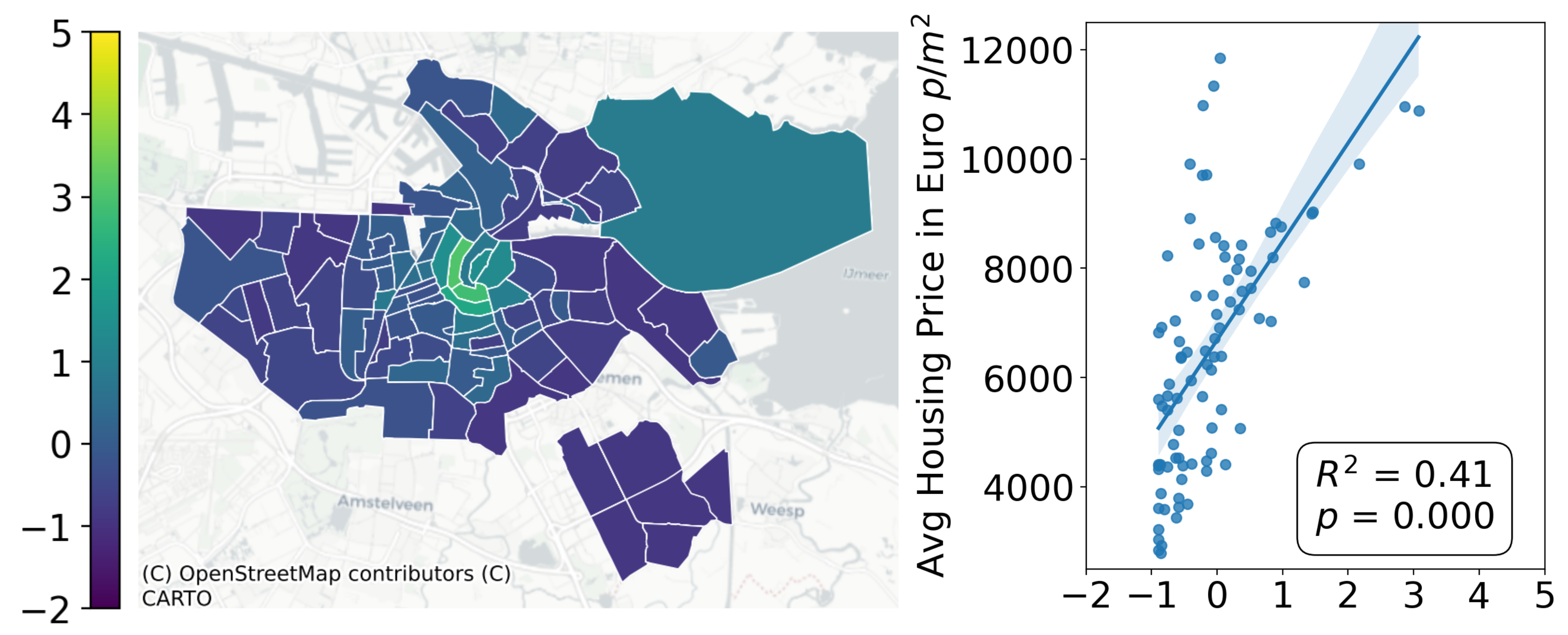}
        \caption{$\sigma$ of Detected Small Changes per Neighbourhoods}
    \end{subfigure}
    \hspace{.1cm}
    \begin{subfigure}{0.49\textwidth}
        \centering
        \includegraphics[width=\textwidth]{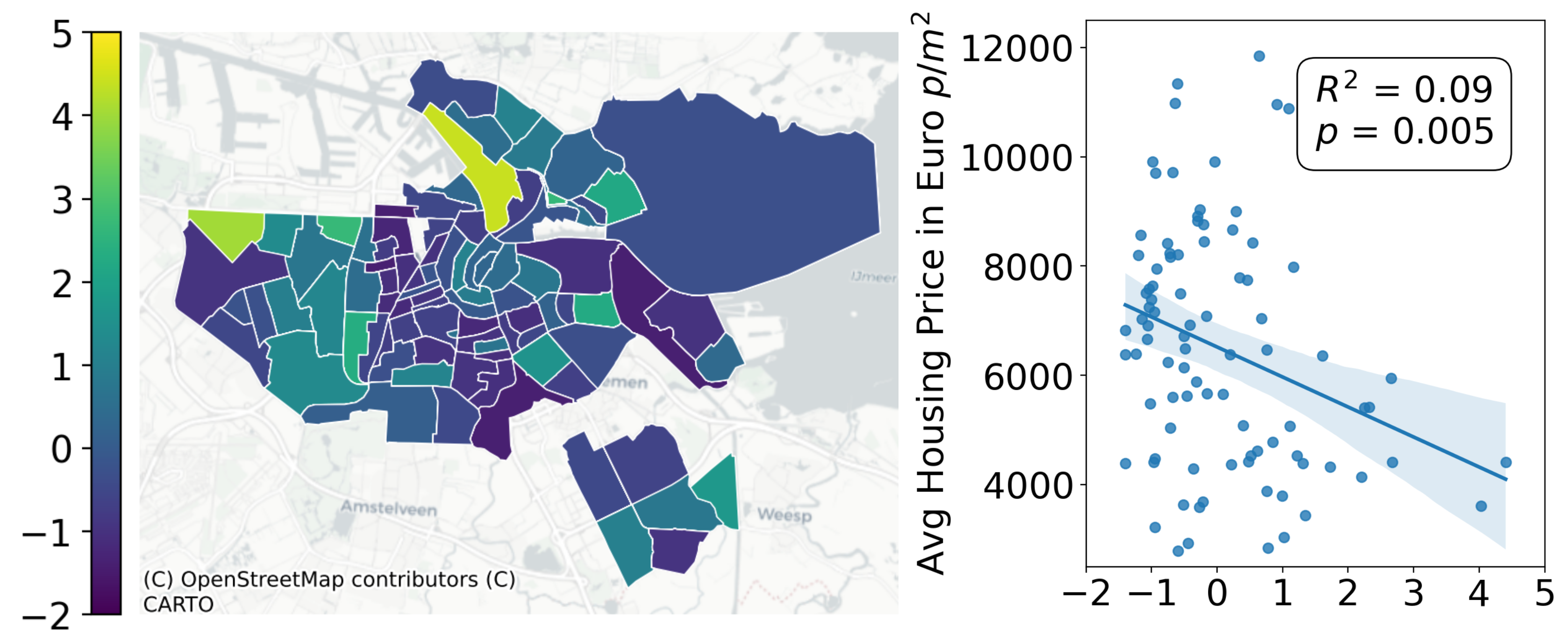}
                \caption{$\sigma$ of Detected Large Changes per Neighbourhoods}
    \end{subfigure}
    \caption{Detected small and large changes per neighbourhood. On the left the small changes are concentrated in the city center and show a positive correlation with housing prices. On the right the large changes are mostly found in the outskirts around large building projects in the North and West, showing a negative trend when plotted against housing prices.}
    \label{fig:casestudycorr}
\end{figure*}

\section{Case Study}
In addition to model comparison we also perform a case study to determine what visual elements of change exists throughout the full urban landscape of Amsterdam. We show how we can uncover different types of change without a priori definition of what constitutes relevant change. Finally, we distinguish between large and small visual change and show how this type of visual change correlates differently with a socio-economic indicator in the form of housing prices, illustrating how citizens of varying socio-economic backgrounds are potentially influenced by change in different ways.  

\subsection{Change Detection Setup}
We run EMPLACE SI-2 on the entire Amsterdam test set and perform zero-shot change detection on the heatmaps on all pairwise comparisons in each cluster. The detection threshold above which we consider the token distance to constitute change is taken from the zero-shot experiments on AMS-buildings and AMS-Trees: We use the window size that performed best for the zero-shot runs for SI-2 in Table~\ref{tab:CityPulseEval}, which is 8x8. Of the 10 runs for which the zero-shot score was calculated, we take the maximum threshold value. We perform pairwise comparisons on all clusters in the AC-1M test set, and restrict our detections to 1 per image pair. This results in $889$ detections on the 25k clusters of the AC-1M test set.  

For small visual change, we cannot simply reduce the window size as the detection could then also be part of a larger change. As such we consider small visual change to be captured in 2x2 tokens, and not spill over to other neighbouring tokens. To capture these small changes, we perform a convolution operation to ensure the average of the tokens in a window is above the threshold value, while also being at least 120\% larger than each of the surrounding tokens. Note that due to perpendicularity the tokens on the left and right side of the images wrap around. This results in $35177$ detections on the 25k clusters of the AC-1M test set. Examples of retrieved small and large changes can be found in the Appendix in Figure~A.2 and Figure~A.3 respectively.

\subsection{Correlation with Socio-Economic Indicators}
To link the found changes to a socio-economic indicator we use the housing price dataset of \cite{groenen2022panorams}. The housing prices are available in a bin system for each ``wijk'' a cluster of neighbourhoods, which gives us 98 datapoints. We aggregate the changes per neighbourhood and show them in Figure~\ref{fig:casestudycorr}. We observe that for small changes, there is a positive correlation with housing prices with an $R^2$ of $.41$ and a $p$ value near 0. For large changes, we observe a negative trend. Because the $R^2$ is $0.09$, we can not point to a negative correlation. However, the fact that the same correlation that holds for small changes does not hold for large changes points to the fact citizens of neighbourhoods with differing socio-economic demographics are confronted with different types of visual change. While small change happens mostly in the center of the city, large changes exist in the outskirts, where housing projects are being built on a continuous basis.

\section{Conclusion}
Our goal was to detect change throughout the city of Amsterdam. We built three datasets including AC-1M, the large USCD dataset to date, and built EMPLACE, a self-supervised method to learn strong change detection features. We introduced a triplet loss, cut-and-flip data augmentation, and an evaluation method in the form of order prediction and  showed that EMPLACE was able to outperform SOTA methods for discrete change detection as a pre-training method for both buildings as well as trees. Additionally, we showed that EMPLACE was able to generate more distinct change heatmaps and that zero-shot change prediction scored high enough to detect various urban changes within Amsterdam. 

In addition, we performed change detection on the entirety of Amsterdam, uncovered large and small visual changes, and showed that, in Amsterdam, expensive neighbourhoods are more likely to experience small visual change, while the inverse is true for large visual changes. We believe our work paves the way for a direction of USCD, and VUA by extent, that tries to uncover visual changes without an a priori definition of what to look for and can therefore be used more directly to investigate urban environments.  

\bibliography{aaai25}

\begin{thebibliography}{34}
\providecommand{\natexlab}[1]{#1}

\bibitem[{Alcantarilla et~al.(2016)Alcantarilla, Stent, Ros, Arroyo, and
  Gherardi}]{Alcantarilla2016}
Alcantarilla, P.~F.; Stent, S.; Ros, G.; Arroyo, R.; and Gherardi, R. 2016.
\newblock {Street-view change detection with deconvolutional networks}.
\newblock \emph{Robotics: Science and Systems}, 12.

\bibitem[{Alpherts, van Noord, and Ghebreab(2023)}]{Alpherts_panoparadigm_2023}
Alpherts, T.; van Noord, N.; and Ghebreab, S. 2023.
\newblock {Explaining Neighbourhood Liveability with Computer Vision: A
  Comparison of Methods on Usability and Practicality}.
\newblock Technical report.

\bibitem[{Batty(2019)}]{Batty2019}
Batty, M. 2019.
\newblock {Urban analytics defined}.
\newblock \emph{Environment and Planning B: Urban Analytics and City Science},
  46(3): 403--405.

\bibitem[{Chen, Yang, and Stiefelhagen(2021)}]{chen2021drtanet}
Chen, S.; Yang, K.; and Stiefelhagen, R. 2021.
\newblock DR-TANet: Dynamic Receptive Temporal Attention Network for Street
  Scene Change Detection.
\newblock arXiv:2103.00879.

\bibitem[{Dubey et~al.(2016)Dubey, Nikhil, Parikh, Raskar, and
  Hidalgo}]{Dubey2016}
Dubey, A.; Nikhil, N.; Parikh, D.; Raskar, R.; and Hidalgo, C.~A. 2016.
\newblock {Deep Learning the City: Quantifying Urban Perception at a Global
  Scale}.
\newblock \emph{Eccv}, 3: 398--413.

\bibitem[{Ester et~al.(1996)Ester, Kriegel, Sander, and Xu}]{dbscan}
Ester, M.; Kriegel, H.-P.; Sander, J.; and Xu, X. 1996.
\newblock A density-based algorithm for discovering clusters in large spatial
  databases with noise.
\newblock In \emph{Proceedings of the Second International Conference on
  Knowledge Discovery and Data Mining}, KDD'96, 226–231. AAAI Press.

\bibitem[{Gebru et~al.(2017)Gebru, Krause, Wang, Chen, Deng, Aiden, and
  Fei-Fei}]{Gebru2017}
Gebru, T.; Krause, J.; Wang, Y.; Chen, D.; Deng, J.; Aiden, E.~L.; and Fei-Fei,
  L. 2017.
\newblock {Using deep learning and google street view to estimate the
  demographic makeup of neighborhoods across the United States}.
\newblock \emph{Proceedings of the National Academy of Sciences of the United
  States of America}, 114(50): 13108--13113.

\bibitem[{Groenen, Rudinac, and Worring(2022)}]{groenen2022panorams}
Groenen, I.; Rudinac, S.; and Worring, M. 2022.
\newblock PanorAMS: Automatic Annotation for Detecting Objects in Urban
  Context.
\newblock arXiv:2208.14295.

\bibitem[{He et~al.(2015)He, Zhang, Ren, and
  Sun}]{he2015deepresiduallearningimage}
He, K.; Zhang, X.; Ren, S.; and Sun, J. 2015.
\newblock Deep Residual Learning for Image Recognition.
\newblock arXiv:1512.03385.

\bibitem[{Huang et~al.(2024)Huang, Wu, Wu, Hwang, and
  Rajagopal}]{huang2024citypulse}
Huang, T.; Wu, Z.; Wu, J.; Hwang, J.; and Rajagopal, R. 2024.
\newblock CityPulse: Fine-Grained Assessment of Urban Change with Street View
  Time Series.
\newblock arXiv:2401.01107.

\bibitem[{Ibrahimi et~al.(2021)Ibrahimi, van Noord, Alpherts, and
  Worring}]{ibrahimi2021inside}
Ibrahimi, S.; van Noord, N.; Alpherts, T.; and Worring, M. 2021.
\newblock Inside Out Visual Place Recognition.
\newblock arXiv:2111.13546.

\bibitem[{Joglekar et~al.(2020)Joglekar, Quercia, Redi, Aiello, Kauer, and
  Sastry}]{Joglekar2020}
Joglekar, S.; Quercia, D.; Redi, M.; Aiello, L.~M.; Kauer, T.; and Sastry, N.
  2020.
\newblock {Facelift: A transparent deep learning framework to beautify urban
  scenes}.
\newblock \emph{Royal Society Open Science}, 7(1).

\bibitem[{Koch and Brilakis(2011)}]{Koch2011}
Koch, C.; and Brilakis, I. 2011.
\newblock {Pothole detection in asphalt pavement images}.
\newblock \emph{Advanced Engineering Informatics}, 25(3): 507--515.

\bibitem[{Law, Paige, and Russell(2019)}]{Law2019}
Law, S.; Paige, B.; and Russell, C. 2019.
\newblock {Take a look around: Using street view and satellite images to
  estimate house prices}.
\newblock \emph{ACM Transactions on Intelligent Systems and Technology}, 10(5):
  1--19.

\bibitem[{Lei et~al.(2021)Lei, Peng, Zhang, Ke, and Li}]{Lei2021}
Lei, Y.; Peng, D.; Zhang, P.; Ke, Q.; and Li, H. 2021.
\newblock Hierarchical Paired Channel Fusion Network for Street Scene Change
  Detection.
\newblock \emph{{IEEE} Transactions on Image Processing}, 30: 55--67.

\bibitem[{Li et~al.(2015)Li, Zhang, Li, Kuzovkina, and Weiner}]{Li2015}
Li, X.; Zhang, C.; Li, W.; Kuzovkina, Y.~A.; and Weiner, D. 2015.
\newblock Who lives in greener neighborhoods? The distribution of street
  greenery and its association with residents’ socioeconomic conditions in
  Hartford, Connecticut, USA.
\newblock \emph{Urban Forestry \& Urban Greening}, 14(4): 751--759.

\bibitem[{Ma et~al.(2022)Ma, Fan, Wang, Wu, Jiang, Xie, and Fan}]{Ma2022}
Ma, N.; Fan, J.; Wang, W.; Wu, J.; Jiang, Y.; Xie, L.; and Fan, R. 2022.
\newblock {Computer Vision for Road Imaging and Pothole Detection: A
  State-of-the-Art Review of Systems and Algorithms}.
\newblock 1--16.

\bibitem[{Muller et~al.(2022)Muller, Gemmell, Choudhury, Nathvani, Metzler,
  Bennett, Denton, Flaxman, and Ezzati}]{Muller2022}
Muller, E.; Gemmell, E.; Choudhury, I.; Nathvani, R.; Metzler, A.~B.; Bennett,
  J.; Denton, E.; Flaxman, S.; and Ezzati, M. 2022.
\newblock \emph{{City-Wide Perceptions of Neighbourhood Quality using Street
  View Images}}, volume~1.
\newblock Association for Computing Machinery.

\bibitem[{Naik et~al.(2017)Naik, Kominers, Raskar, Glaeser, and
  Hidalgo}]{Naik2017}
Naik, N.; Kominers, S.~D.; Raskar, R.; Glaeser, E.~L.; and Hidalgo, C.~A. 2017.
\newblock {Computer vision uncovers predictors of physical urban change}.
\newblock \emph{Proceedings of the National Academy of Sciences of the United
  States of America}, 114(29): 7571--7576.

\bibitem[{Naik et~al.(2014)Naik, Philipoom, Raskar, and Hidalgo}]{Naik2014}
Naik, N.; Philipoom, J.; Raskar, R.; and Hidalgo, C. 2014.
\newblock {Streetscore-predicting the perceived safety of one million
  streetscapes}.
\newblock \emph{IEEE Computer Society Conference on Computer Vision and Pattern
  Recognition Workshops}, (January): 793--799.

\bibitem[{Oquab et~al.(2024)Oquab, Darcet, Moutakanni, Vo, Szafraniec,
  Khalidov, Fernandez, Haziza, Massa, El-Nouby, Assran, Ballas, Galuba, Howes,
  Huang, Li, Misra, Rabbat, Sharma, Synnaeve, Xu, Jegou, Mairal, Labatut,
  Joulin, and Bojanowski}]{oquab2024dinov2}
Oquab, M.; Darcet, T.; Moutakanni, T.; Vo, H.; Szafraniec, M.; Khalidov, V.;
  Fernandez, P.; Haziza, D.; Massa, F.; El-Nouby, A.; Assran, M.; Ballas, N.;
  Galuba, W.; Howes, R.; Huang, P.-Y.; Li, S.-W.; Misra, I.; Rabbat, M.;
  Sharma, V.; Synnaeve, G.; Xu, H.; Jegou, H.; Mairal, J.; Labatut, P.; Joulin,
  A.; and Bojanowski, P. 2024.
\newblock DINOv2: Learning Robust Visual Features without Supervision.
\newblock arXiv:2304.07193.

\bibitem[{Ordonez and Berg(2014)}]{Ordonez2014}
Ordonez, V.; and Berg, T.~L. 2014.
\newblock {Learning high-level judgments of urban perception}.
\newblock \emph{Lecture Notes in Computer Science (including subseries Lecture
  Notes in Artificial Intelligence and Lecture Notes in Bioinformatics)}, 8694
  LNCS(PART 6): 494--510.

\bibitem[{Radford et~al.(2021)Radford, Kim, Hallacy, Ramesh, Goh, Agarwal,
  Sastry, Askell, Mishkin, Clark, Krueger, and
  Sutskever}]{radford2021learningtransferablevisualmodels}
Radford, A.; Kim, J.~W.; Hallacy, C.; Ramesh, A.; Goh, G.; Agarwal, S.; Sastry,
  G.; Askell, A.; Mishkin, P.; Clark, J.; Krueger, G.; and Sutskever, I. 2021.
\newblock Learning Transferable Visual Models From Natural Language
  Supervision.
\newblock arXiv:2103.00020.

\bibitem[{Ramkumar, Arani, and Zonooz(2022)}]{Ramkumar2022}
Ramkumar, V. R.~T.; Arani, E.; and Zonooz, B. 2022.
\newblock {Differencing based Self-supervised pretraining for Scene Change
  Detection}.
\newblock (11): 1--13.

\bibitem[{Sakurada(2018)}]{Sakurada2018}
Sakurada, K. 2018.
\newblock {Weakly Supervised Silhouette-based Semantic Change Detection}.

\bibitem[{Sakurada and Okatani(2015)}]{Sakurada2015}
Sakurada, K.; and Okatani, T. 2015.
\newblock {Change Detection from a Street Image Pair using CNN Features and
  Superpixel Segmentation}.
\newblock 61.1--61.12.

\bibitem[{Seiferling et~al.(2017)Seiferling, Naik, Ratti, and
  Proulx}]{Seiferling2017}
Seiferling, I.; Naik, N.; Ratti, C.; and Proulx, R. 2017.
\newblock {Green streets: Quantifying and mapping urban trees with street-level
  imagery and computer vision}.
\newblock \emph{Landscape and Urban Planning}, 165(4): 93--101.

\bibitem[{Seresinhe, Preis, and Moat(2017)}]{Seresinhe2017}
Seresinhe, C.~I.; Preis, T.; and Moat, H.~S. 2017.
\newblock {Using deep learning to quantify the beauty of outdoor places}.
\newblock \emph{Royal Society Open Science}, 4(7).

\bibitem[{Suel et~al.(2021)Suel, Bhatt, Brauer, Flaxman, and Ezzati}]{Suel2021}
Suel, E.; Bhatt, S.; Brauer, M.; Flaxman, S.; and Ezzati, M. 2021.
\newblock {Multimodal deep learning from satellite and street-level imagery for
  measuring income, overcrowding, and environmental deprivation in urban
  areas}.
\newblock \emph{Remote Sensing of Environment}, 257(June 2020): 112339.

\bibitem[{Suel et~al.(2019)Suel, Polak, Bennett, and Ezzati}]{Suel2019}
Suel, E.; Polak, J.~W.; Bennett, J.~E.; and Ezzati, M. 2019.
\newblock {Measuring social, environmental and health inequalities using deep
  learning and street imagery}.
\newblock \emph{Scientific Reports}, 9(1): 1--10.

\bibitem[{Sukel, Rudinac, and Worring(2020)}]{Sukel2020}
Sukel, M.; Rudinac, S.; and Worring, M. 2020.
\newblock {Urban object detection kit: A system for collection and analysis of
  street-level imagery}.
\newblock \emph{ICMR 2020 - Proceedings of the 2020 International Conference on
  Multimedia Retrieval}, 509--516.

\bibitem[{Varghese et~al.(2018)Varghese, Gubbi, Ramaswamy, and
  Balamuralidhar}]{Varghese2018}
Varghese, A.; Gubbi, J.; Ramaswamy, A.; and Balamuralidhar, P. 2018.
\newblock {ChangeNet: A Deep Learning Architecture for Visual Change
  Detection.pdf}.

\bibitem[{Yildiz et~al.(2022)Yildiz, Khademi, Siebes, and van
  Gemert}]{yildiz2022amstertime}
Yildiz, B.; Khademi, S.; Siebes, R.~M.; and van Gemert, J. 2022.
\newblock AmsterTime: A Visual Place Recognition Benchmark Dataset for Severe
  Domain Shift.
\newblock arXiv:2203.16291.

\bibitem[{Zhao et~al.(2019)Zhao, Li, Wang, Zheng, and Shi}]{Zhao2019}
Zhao, X.; Li, H.; Wang, R.; Zheng, C.; and Shi, S. 2019.
\newblock {Street-view Change Detection via Siamese Encoder-decoder Structured
  Convolutional Neural Networks}.
\newblock \emph{VISIGRAPP 2019 - Proceedings of the 14th International Joint
  Conference on Computer Vision, Imaging and Computer Graphics Theory and
  Applications}, 5(Visigrapp): 525--532.

\end{thebibliography}

\end{document}

\maketitle

\begin{table}
    \centering
    \begin{tabular}{c|c|cc|cc|cc|cc|cc}
    
    \multirow{2}{*}{Cut-and-Flip} & \multirow{2}{*}{Loss} & \multicolumn{2}{c|}{SI-1} & \multicolumn{2}{c|}{SI-2} & \multicolumn{2}{c|}{SI-3} & \multicolumn{2}{c|}{SI-4} &\multicolumn{2}{c}{SI-Hard} \\
    \cline{3-12}
                                  &                       & Acc & Epoch & Acc & Epoch & Acc & Epoch & Acc & Epoch & Acc & Epoch \\
    \hline
    \cmark                        & Adaptive              & $.975$ & $13$ & $.902$ & $11$ & $.937$ & $15$ & $.951$ & $13$ & $.928$ & $16$\\
    \xmark                        & Adaptive              & $.965$ & $14$   & $.870$ & $17$   & $.926$ & $14$   & $.927$ & $12$ & $.888$ & $21$  \\
    %\xmark                        & Fixed                 & $-$ & $-$   & $-$ & $-$   & $-$ & $-$   & $-$ & $-$  & $-$ & $-$ \\
    \cmark                        & Fixed                 & $.954$ & $18$ & $.875$ & $26$ & $.929$ & $17$ & $.948$ & $17$& $.908$ & $22$ \\
    \hline
    \end{tabular}
    \caption{Ablation of cut-and-flip augmentation and adaptive triplet loss. Fixed margin set to $\alpha = 1$. SI-Hard parameters are $90<\Delta_{AP}<365$ and $365<\Delta_{AN}$.}
    \label{tab:ablation}
\end{table}

\section{Appendix}

\subsection{Ablation Study}
We evaluate the utility of both cut-and-flip augmentation as well as the adaptive triplet loss introduced in Eq. 6. The results are shown in Table~\ref{tab:ablation}. We see that cut-and-flip augmentation provides an increase in performance over all setups. We also see that the adaptive margin has increased utility on setups where the negative and positive are closer together such as SI-2. We also observe that in all instances where the model trains, the adaptive margin takes the fewest number of epochs to converge. As such we believe the temporal triplet loss outperforms the fixed margin by performing similar or better in all tested setups.
\begin{table}[H]
    \centering
    \begin{tabular}{l|c|c|c|c|c}
        & SI-1 & SI-2 & SI-3 & SI-4 & All \\
        \hline
        SI-1 & \textbf{.013} & .040 & .065 & .099 & .055 \\
        SI-2 & .014 & \textbf{.022} & \textbf{.024} & \textbf{.010} & \textbf{.015} \\
        SI-3 & .031 & .037 & .026 & .017 & .018 \\
        SI-4 & .030 & .048 & .037 & .033 & .041 \\
        Baseline & .081 & .028 & .054 & .083 & .510 \\
        \hline
    \end{tabular}
    \caption{Standard deviation over the accuracy distribution over the 8 geographical areas (stadsdelen) of Amsterdam.}
    \label{tab:biasstudy}
\end{table}

\subsection*{\phantom{}}
\vspace{2cm} % Adjust as needed
\subsection{Bias study}

To enable learning in a self-supervised manner we build on the assumption that temporal change is proportional to visual/urban change. However, this assumption may not strictly hold in every case, as there may be certain places in the city that hardly change over the years. As such, this can potentially cause our model to be biased towards places that exhibit more visual change. To study possible bias as a result of this assumption we repeat the order prediction experiment in Table 3 split by the 8 geographical areas (stadsdelen) of Amsterdam. The values in this Table represent the standard deviation of the distribution of accuracies over the spatial regions. While the variance between the accuracies over different regions is present, we see that overall this variance is larger in easier setups. Harder setups, such as SI-2, show less variance over the different regions. From this we conclude that baseline methods show strong spatial biases and that training with EMPLACE shows a significant reduction in this bias.

\begin{figure*}
    \centering
    \includegraphics[width=0.8\linewidth]{CameraReady/LaTeX/Images/cluster_plot_lightseagreen.png}
    \caption{Plot of the 254k clusters of the AC-1M.}
    \label{fig:appendix_clusters_on_map_of_amsterdam}
\end{figure*}

\begin{figure*}
\centering
\caption{Detected small changes in Amsterdam with zero-shot EMPLACE. For each set of images, change was detected between image 2 and 3.}
\setlength{\tabcolsep}{5pt}
\renewcommand{\arraystretch}{1.5}
\begin{tabular}{ccc}
%%% Kleurio
\includegraphics[width=.3\textwidth]{CameraReady/LaTeX/Changes/calclin22/Aalsmeerwegbuurt_Oost_20210219_pid_0000_001555_20210420_pid_0003_000108_20220322_pid_0000_000135_20220324_pid_0001_000196_5.jpg}&
\includegraphics[width=.3\textwidth]{CameraReady/LaTeX/Changes/calclin22/Venserpolder_West_20160804_pid_0000_003537_20170615_pid_0000_010483_20200630_pid_0000_002470_20220620_pid_0005_001828_1.jpg}&
\includegraphics[width=.3\textwidth]{CameraReady/LaTeX/Changes/calclin22/Werengouw_Zuid_20160805_pid_0000_006527_20180725_pid_0000_003751_20190708_pid_0000_001209_20210615_pid_0000_002902_10.jpg}\\
%%% Winkels
\includegraphics[width=.3\textwidth]{CameraReady/LaTeX/Changes/calclin22/Borgerbuurt_20190109_pid_0002_002754_20190430_pid_0000_001043_20200511_pid_0000_005695_20210325_pid_0000_006304_0.jpg}&
\includegraphics[width=.3\textwidth]{CameraReady/LaTeX/Changes/calclin22/Trompbuurt_20200508_pid_0001_004693_20210222_pid_0000_001669_20220502_pid_0000_000553_20221214_pid_0003_000009_0.jpg}&
\includegraphics[width=.3\textwidth]{CameraReady/LaTeX/Changes/calclin22/Tuindorp_Oostzaan_Oost_20160622_pid_0001_001310_20160706_pid_0000_001128_20190705_pid_0000_002137_20210624_pid_0000_004137_8.jpg}\\

\includegraphics[width=.3\textwidth]{CameraReady/LaTeX/Changes/calclin22/Uilenburg_20210318_pid_0000_000020_20210915_pid_0000_000321_20220425_pid_0000_000264_20220610_pid_0000_000021_2.jpg}&
\includegraphics[width=.3\textwidth]{CameraReady/LaTeX/Changes/calclin22/Vondelparkbuurt_Midden_20180612_pid_0003_000626_20190107_pid_0002_001686_20210222_pid_0000_000465_20220902_pid_0000_000577_0.jpg}&
\includegraphics[width=.3\textwidth]{CameraReady/LaTeX/Changes/calclin22/Scheepvaarthuisbuurt_20190114_pid_0002_002553_20190114_pid_0002_002557_20210308_pid_0001_000214_20220426_pid_0002_000698_19.jpg}\\
%%% Textielbakken

\includegraphics[width=.3\textwidth]{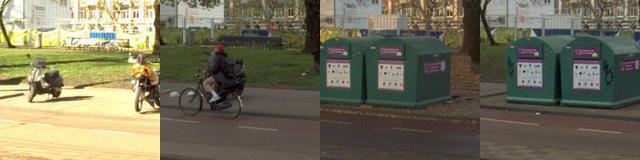}&
\includegraphics[width=.3\textwidth]{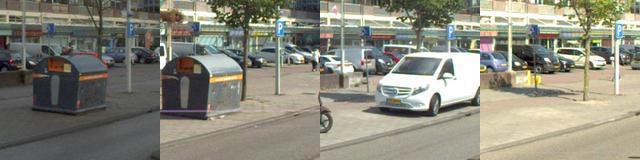}&
\includegraphics[width=.3\textwidth]{CameraReady/LaTeX/Changes/calclin22/Rijnbuurt_West_20170407_pid_0000_002632_20180605_pid_0000_001331_20200901_pid_0001_000331_20200917_pid_0000_000900_0.jpg}\\
%Bloembakken op straat
\includegraphics[width=.3\textwidth]{CameraReady/LaTeX/Changes/calclin22/Marathonbuurt_West_20160727_pid_0002_002031_20190503_pid_0000_003970_20210219_pid_0000_001097_20210420_pid_0008_000913_0.jpg}&
\includegraphics[width=.3\textwidth]{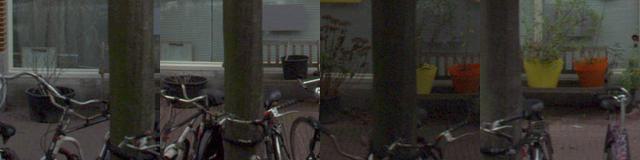}&
\includegraphics[width=.3\textwidth]{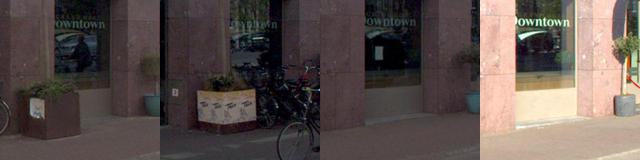}\\

%Nieuw Stoepie
\includegraphics[width=.3\textwidth]{CameraReady/LaTeX/Changes/calclin22/Bedrijventerrein_Schinkel_20180612_pid_0001_002429_20181115_pid_0001_002429_20191204_pid_0000_000171_20200506_pid_0000_000402_0.jpg}&
\includegraphics[width=.3\textwidth]{CameraReady/LaTeX/Changes/calclin22/Bloemgrachtbuurt_20160815_pid_0007_001117_20160817_pid_0002_000049_20210317_pid_0001_000285_20220425_pid_0006_000181_56.jpg}&
\includegraphics[width=.3\textwidth]{CameraReady/LaTeX/Changes/calclin22/Banne_Zuidwest_20180802_pid_0004_000322_20190702_pid_0001_001346_20210623_pid_0000_007877_20210625_pid_0000_000036_14.jpg}\\

%Nieuw Stoepie
\includegraphics[width=.3\textwidth]{CameraReady/LaTeX/Changes/calclin22/Blauwe_Zand_20190627_pid_0000_000103_20200528_pid_0000_001278_20200916_pid_0000_000728_20210617_pid_0000_001055_10.jpg}&
\includegraphics[width=.3\textwidth]{CameraReady/LaTeX/Changes/calclin22/Tuindorp_Oostzaan_Oost_20160622_pid_0001_001311_20160706_pid_0000_001129_20190705_pid_0000_002138_20210624_pid_0000_004138_8.jpg}&
\includegraphics[width=.3\textwidth]{CameraReady/LaTeX/Changes/calclin22/Banne_Zuidwest_20180802_pid_0004_000322_20190702_pid_0001_001346_20210623_pid_0000_007877_20210625_pid_0000_000036_12.jpg}\\
%Nieuw Wegdek
\includegraphics[width=.3\textwidth]{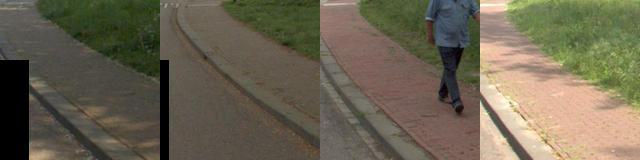}&
\includegraphics[width=.3\textwidth]{CameraReady/LaTeX/Changes/calclin22/Begijnhofbuurt_20190115_pid_0000_000844_20190115_pid_0000_000873_20200421_pid_0000_002532_20200424_pid_0000_000063_66.jpg}&
\includegraphics[width=.3\textwidth]{CameraReady/LaTeX/Changes/calclin22/Blauwe_Zand_20190117_pid_0000_000543_20190627_pid_0002_001806_20210615_pid_0000_003276_20220512_pid_0001_001320_3.jpg}\\

%WIOR
\includegraphics[width=.3\textwidth]{CameraReady/LaTeX/Changes/calclin22/Osdorpplein_eo_20200821_pid_0000_001915_20200824_pid_0000_004148_20210222_pid_0000_000745_20210709_pid_0000_000091_2.jpg}&
\includegraphics[width=.3\textwidth]{CameraReady/LaTeX/Changes/calclin22/Meer_en_Oever_20190717_pid_0000_007450_20200824_pid_0000_004187_20210709_pid_0000_000130_20220411_pid_0003_001262_0.jpg}&
\includegraphics[width=.3\textwidth]{CameraReady/LaTeX/Changes/calclin22/G-buurt_West_20170615_pid_0000_016106_20180712_pid_0000_002161_20200703_pid_0000_002416_20210603_pid_0000_004431_1.jpg}\\

\includegraphics[width=.3\textwidth]{CameraReady/LaTeX/Changes/calclin22/Blauwe_Zand_20170515_pid_0000_003160_20180730_pid_0001_000874_20200206_pid_0007_000045_20200206_pid_0007_000046_101.jpg}&
\includegraphics[width=.3\textwidth]{CameraReady/LaTeX/Changes/calclin22/Buitenveldert_Zuidoost_20190506_pid_0000_005038_20200706_pid_0000_005531_20210506_pid_0000_001231_20210507_pid_0000_002866_8.jpg}&
\includegraphics[width=.3\textwidth]{CameraReady/LaTeX/Changes/calclin22/Langestraat_eo_20210317_pid_0001_001489_20210915_pid_0000_001824_20221231_pid_0012_000010_20230130_pid_0000_000088_9.jpg}\\

%Hek of Schutting
\includegraphics[width=.3\textwidth]{CameraReady/LaTeX/Changes/calclin22/Buitenveldert_Zuidwest_20180530_pid_0000_005101_20190507_pid_0001_003611_20200828_pid_0000_000871_20210506_pid_0000_002200_3.jpg}&
\includegraphics[width=.3\textwidth]{CameraReady/LaTeX/Changes/calclin22/Terrasdorp_20180803_pid_0001_000209_20200616_pid_0000_005445_20210625_pid_0000_001312_20220510_pid_0000_002767_1.jpg}&
\includegraphics[width=.3\textwidth]{CameraReady/LaTeX/Changes/calclin22/Van_der_Pekbuurt_20191212_pid_0002_000626_20200603_pid_0000_002029_20210616_pid_0000_001739_20220512_pid_0006_005877_0.jpg}\\

%%Groen op gebouwtj
\includegraphics[width=.3\textwidth]{CameraReady/LaTeX/Changes/calclin22/Paramariboplein_eo_20170420_pid_0000_000181_20190426_pid_0000_001094_20200511_pid_0000_003750_20210325_pid_0000_000912_1.jpg}&
\includegraphics[width=.3\textwidth]{CameraReady/LaTeX/Changes/calclin22/Weesperzijde_Midden-Zuid_20160517_pid_0001_001675_20170331_pid_0002_000355_20181212_pid_0006_000292_20200904_pid_0002_000941_11.jpg}&
\includegraphics[width=.3\textwidth]{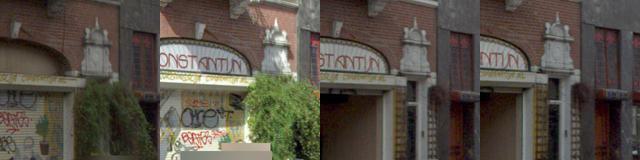}\\
\end{tabular}
\label{appendix_small_changes}
\end{figure*}

\begin{figure*}
\centering
\caption{Detected large changes in Amsterdam with zero-shot EMPLACE.}
\setlength{\tabcolsep}{5pt}
\renewcommand{\arraystretch}{1.5}
\begin{tabular}{cccc}
%% Nieuw Gebouw - GROOT
\includegraphics[width=.225\textwidth]{CameraReady/LaTeX/Changes/88/Swammerdambuurt_20191219_pid_000499_20230102_pid_000114_234_55-302.jpg}&
\includegraphics[width=.225\textwidth]{CameraReady/LaTeX/Changes/88/Amsterdamse_Poort_20160707_pid_007706_20190613_pid_003123_201_50-170.jpg}&
\includegraphics[width=.225\textwidth]{CameraReady/LaTeX/Changes/88/Tuindorp_Amstelstation_20160419_pid_000544_20210507_pid_001572_272_50-203.jpg}&
\includegraphics[width=.225\textwidth]{CameraReady/LaTeX/Changes/88/Valeriusbuurt_Oost_20190117_pid_000840_20230110_pid_000180_281_51-523.jpg}\\
%% Nieuw huisje
\includegraphics[width=.225\textwidth]{CameraReady/LaTeX/Changes/88/Oostzanerdijk_20200206_pid_000110_20221227_pid_000370_215_52-207.jpg}&
\includegraphics[width=.225\textwidth]{CameraReady/LaTeX/Changes/88/Bosleeuw_20160615_pid_000446_20180620_pid_002174_317_50-467.jpg}&
\includegraphics[width=.225\textwidth]{CameraReady/LaTeX/Changes/88/Kortvoort_20180716_pid_003875_20200703_pid_001745_260_50-405.jpg}&
\includegraphics[width=.225\textwidth]{CameraReady/LaTeX/Changes/88/De_Bongerd_20181219_pid_000869_20221227_pid_000052_289_50-062.jpg}\\
%% Grote ingrepen
\includegraphics[width=.225\textwidth]{CameraReady/LaTeX/Changes/88/IJsbaanpad_eo_20170413_pid_000585_20190430_pid_000740_206_50-192.jpg}&
\includegraphics[width=.225\textwidth]{CameraReady/LaTeX/Changes/88/Markengouw_Midden_20181219_pid_000396_20191211_pid_000106_245_51-337.jpg}&
\includegraphics[width=.225\textwidth]{CameraReady/LaTeX/Changes/88/De_Eenhoorn_20170406_pid_000730_20180524_pid_000188_303_50-440.jpg}&
\includegraphics[width=.225\textwidth]{CameraReady/LaTeX/Changes/88/Venserpolder_Oost_20170621_pid_000658_20210604_pid_003901_213_57-498.jpg}\\
%% Steiger met zeil erbij
\includegraphics[width=.225\textwidth]{CameraReady/LaTeX/Changes/88/Aalsmeerwegbuurt_Oost_20170419_pid_001523_20220324_pid_000396_325_50-211.jpg}&
\includegraphics[width=.225\textwidth]{CameraReady/LaTeX/Changes/88/Aalsmeerwegbuurt_Oost_20190426_pid_002020_20220324_pid_000286_264_50-937.jpg}&
\includegraphics[width=.225\textwidth]{CameraReady/LaTeX/Changes/88/BG-terrein_eo_20190114_pid_001767_20230227_pid_000007_95_50-515.jpg}&
\includegraphics[width=.225\textwidth]{CameraReady/LaTeX/Changes/88/Utrechtsebuurt_Zuid_20190114_pid_000985_20211229_pid_000619_268_50-692.jpg}\\
%% Steiger met zeil van een afstandje
\includegraphics[width=.225\textwidth]{CameraReady/LaTeX/Changes/88/Museumplein_20170412_pid_002231_20220322_pid_001441_325_55-260.jpg}&
\includegraphics[width=.225\textwidth]{CameraReady/LaTeX/Changes/88/Duivelseiland_20220111_pid_000188_20230103_pid_000045_281_55-934.jpg}&
\includegraphics[width=.225\textwidth]{CameraReady/LaTeX/Changes/88/Banpleinbuurt_20180606_pid_002311_20190503_pid_001695_229_58-120.jpg}&
\includegraphics[width=.225\textwidth]{CameraReady/LaTeX/Changes/88/Transvaalbuurt_West_20160608_pid_000050_20190515_pid_003140_124_56-617.jpg}\\
%% Steiger met zeil eraf
\includegraphics[width=.225\textwidth]{CameraReady/LaTeX/Changes/88/Aalsmeerwegbuurt_Oost_20180612_pid_003404_20200827_pid_001331_271_50-215.jpg}&
\includegraphics[width=.225\textwidth]{CameraReady/LaTeX/Changes/88/Aalsmeerwegbuurt_Oost_20180612_pid_003404_20210709_pid_000082_96_50-098.jpg}&
\includegraphics[width=.225\textwidth]{CameraReady/LaTeX/Changes/88/Van_der_Pekbuurt_20211224_pid_000019_20221227_pid_000023_207_50-806.jpg}&
\includegraphics[width=.225\textwidth]{CameraReady/LaTeX/Changes/88/Transvaalbuurt_Oost_20181211_pid_000161_20200129_pid_000458_174_50-292.jpg}\\
%% Steiger zonder zeil
\includegraphics[width=.225\textwidth]{CameraReady/LaTeX/Changes/88/WG_terrein_20160809_pid_001656_20180612_pid_002052_218_50-366.jpg}&
\includegraphics[width=.225\textwidth]{CameraReady/LaTeX/Changes/88/De_Wetbuurt_20210507_pid_003126_20220518_pid_001272_300_55-046.jpg}&
\includegraphics[width=.225\textwidth]{CameraReady/LaTeX/Changes/88/Rode_Kruisbuurt_20160808_pid_002375_20210615_pid_003465_172_50-184.jpg}&
\includegraphics[width=.225\textwidth]{CameraReady/LaTeX/Changes/88/Schinkelbuurt_Zuid_20210219_pid_001306_20230105_pid_000008_125_50-561.jpg}\\
% Nieuw Gebouw closeup
\includegraphics[width=.225\textwidth]{CameraReady/LaTeX/Changes/88/Jeugdland_20181218_pid_002411_20210217_pid_000754_96_50-597.jpg}&
\includegraphics[width=.225\textwidth]{CameraReady/LaTeX/Changes/88/Aalsmeerwegbuurt_Oost_20190107_pid_000474_20220111_pid_000035_170_50-023.jpg}&
\includegraphics[width=.225\textwidth]{CameraReady/LaTeX/Changes/88/Sporenburg_20160517_pid_001633_20200422_pid_003402_93_50-473.jpg}&
\includegraphics[width=.225\textwidth]{CameraReady/LaTeX/Changes/88/Buurt_7_20170519_pid_001360_20210401_pid_003197_135_50-066.jpg}\\
\includegraphics[width=.225\textwidth]{CameraReady/LaTeX/Changes/88/Gein_Noordoost_20190606_pid_001372_20210609_pid_005891_265_50-001.jpg}&
\includegraphics[width=.225\textwidth]{CameraReady/LaTeX/Changes/88/Tuindorp_Frankendael_20180525_pid_001547_20210531_pid_001070_230_50-714.jpg}&
\includegraphics[width=.225\textwidth]{CameraReady/LaTeX/Changes/88/Aalsmeerwegbuurt_Oost_20190107_pid_000100_20230120_pid_000238_335_51-143.jpg}&
\includegraphics[width=.225\textwidth]{CameraReady/LaTeX/Changes/88/G-buurt_Oost_20170615_pid_008011_20210603_pid_005428_220_50-916.jpg}\\
\includegraphics[width=.225\textwidth]{CameraReady/LaTeX/Changes/88/Banne_Zuidoost_20160718_pid_000010_20220513_pid_001422_225_56-429.jpg}&
\includegraphics[width=.225\textwidth]{CameraReady/LaTeX/Changes/88/Nintermanterrein_20190627_pid_003013_20210617_pid_001961_272_50-634.jpg}&
\includegraphics[width=.225\textwidth]{CameraReady/LaTeX/Changes/88/Meer_en_Oever_20160808_pid_001887_20200821_pid_002290_236_50-343.jpg}&
\includegraphics[width=.225\textwidth]{CameraReady/LaTeX/Changes/88/Calandlaan-Lelylaan_20160808_pid_006810_20190723_pid_002629_216_55-077.jpg}\\
\end{tabular}
\label{appendix_large_changes}
\end{figure*}